\documentclass[10pt,twocolumn,letterpaper]{article}

\usepackage{cvpr}
\usepackage{times}
\usepackage{epsfig}
\usepackage{graphicx}
\usepackage{amsmath}
\usepackage{amssymb}
\usepackage{color}

\usepackage{times}
\usepackage{epsfig}
\usepackage{graphicx}
\usepackage{amsmath}
\usepackage{amssymb}
\usepackage{bm}
\usepackage{subfigure}
\usepackage{multirow}%
\usepackage{color}
\usepackage{upgreek,xspace}

\usepackage{helvet}
\usepackage{courier}
\usepackage{bm}
\usepackage{algorithmic}
\usepackage{graphicx}
\usepackage{subfigure}
\usepackage{multirow}%
\usepackage{amsmath, amssymb}
\usepackage{color}
\usepackage{mathrsfs}
\usepackage{amsthm}
\usepackage{epstopdf}
\usepackage{wrapfig}
\usepackage{picinpar}
\usepackage{threeparttable, eucal}
\usepackage{enumitem}
\usepackage{subfigure}
\usepackage[vlined,ruled,linesnumbered]{algorithm2e}
\usepackage{overpic}
\usepackage{bbm}

\usepackage{epstopdf}

\usepackage{url}
\usepackage[usenames,dvipsnames]{xcolor}

\def\bm{{\bf m}}
\def\bn{{\bf n}}
\def\bp{{\bf p}}

\def\bx{{\bf x}}
\def\by{{\bf y}}

\def\0{{\bf 0}}
\def\1{{\bf 1}}

\def\bH{{\bf H}}

\def\mbD{{\mathbb D}}
\def\mbR{{\mathbb R}}
\def\mbZ{{\mathbb Z}}

\def\mb1{{\mathbbm 1}}
\def\etal{\emph{et.al. }}
\def\ie{\emph{i.e. }}
\def\eg{\emph{e.g. }}
\def\etc{\emph{etc }}

\def\K{{\mathcal{K}}}
\def\I{{\mathbf{X}}}
\def\B{{\mathbf{Y}}}
\def\N{{\mathbf{N}}}
\def\U{{\mathbf{U}}}
\def\V{{\mathbf{V}}}
\def\M{{\mathcal{M}}}
\def\vec{{\mbox{vec}}}
\def\bm{{\mathcal{M}}}
\def\F{{F}}

\usepackage[pagebackref=true,breaklinks=true,letterpaper=true,colorlinks,bookmarks=false]{hyperref}

\cvprfinalcopy 
\ifcvprfinal\fi
\begin{document}

\title{From Motion Blur to Motion Flow: a Deep Learning Solution for \\Removing Heterogeneous Motion Blur}
\author{Dong Gong{$^{\dag\ddag}$}, Jie Yang{$^{\ddag}$}, Lingqiao Liu{$^{\ddag\S}$}, Yanning Zhang{$^{\dag}$}, Ian Reid{$^{\ddag\S}$}, \\Chunhua Shen{$^{\ddag\S}$}, Anton van den Hengel{$^{\ddag\S}$}, Qinfeng Shi{$^{\ddag}$}\\%$^\S$ The corresponding authors.
$^{\dag}$School of Computer Science and Engineering, Northwestern Polytechnical University, China\\
$^{\ddag}$School of Computer Science, The University of Adelaide, $^{\S}$Australian Centre for Robotic Vision\\
{\tt\small edgong01@gmail.com;ynzhang@nwpu.edu.cn;} \\
{\tt\small\!\!\!\!\{jie.yang01;lingqiao.liu;ian.reid;chunhua.shen;anton.vandenhengel;javen.shi\}@adelaide.edu.au}
}

\maketitle
\begin{abstract}
Removing pixel-wise heterogeneous motion blur is challenging due to the ill-posed nature of the problem. The predominant solution is to estimate the blur kernel by adding a prior, but the extensive literature on the subject indicates the difficulty in identifying a prior which is suitably informative, and general. Rather than imposing a prior based on theory, we propose instead to learn one from the data. Learning a prior over the latent image would require modeling all possible image content. The critical observation underpinning our approach is thus that learning the motion flow instead allows the model to focus on the cause of the blur, irrespective of the image content. This is a much easier learning task, but it also avoids the iterative process through which latent image priors are typically applied. Our approach directly estimates the motion flow from the blurred image through a fully-convolutional deep neural network (FCN) and recovers the unblurred image from the estimated motion flow. Our FCN is the first universal end-to-end mapping from the blurred image to the dense motion flow. To train the FCN, we simulate motion flows to generate synthetic blurred-image-motion-flow pairs thus avoiding the need for human labeling. Extensive experiments on challenging realistic blurred images demonstrate that the proposed method outperforms the state-of-the-art.

\end{abstract}

\section{Introduction}
Motion blur is ubiquitous in photography,
especially when using light-weight mobile devices, such as cell-phones and on-board cameras.
While there has been a significant progress on image deblurring \cite{fergus2006removing,cho2009fast,xu2010twophase,pan2014text,pan2016dark,gong2016active}, most work focuses on \emph{spatially-uniform} blur.
Some recent methods \cite{whyte2012non,hirsch2011fast,hu2014joint,kim2014segfree,pan2016soft} have been proposed to remove  \emph{spatially-varying} blur caused by camera panning, and/or object movement, with some restrictive assumptions on the types of blur, image prior, or both. In this work, we focus on recovering
a blur-free latent image from a single observation
degraded by \emph{heterogeneous motion blur}, \ie the blur kernels may independently vary from pixel to pixel.

\begin{figure}[!t]
\centering
\subfigure[Blurry image]{
\begin{minipage}[b]{.23\textwidth}
\centerline{
\begin{overpic}[width=1\textwidth]
{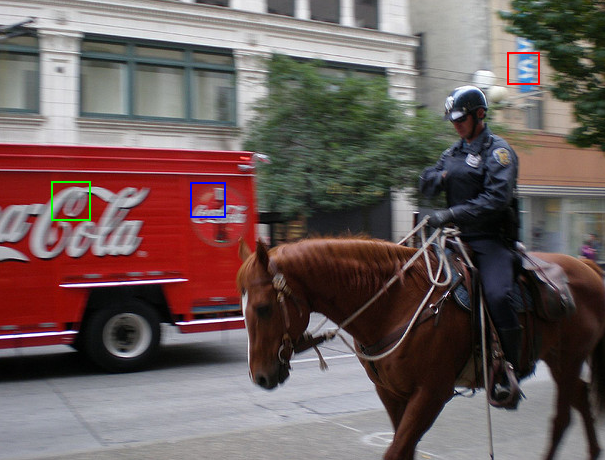}
\put(-2,0.05){ \includegraphics[width=0.23\textwidth]{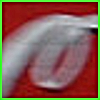}}
\put(21.15,0.05){ \includegraphics[width=0.23\textwidth]{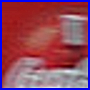}}
\put(44.35,0.05){ \includegraphics[width=0.23\textwidth]{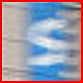}}
\end{overpic}}
\end{minipage}
}
\vspace{-0.2cm}
\hspace{-0.3cm}
\subfigure[Xu and Jia \cite{xu2010twophase}]{
\begin{minipage}[b]{.23\textwidth}
\centerline{
\begin{overpic}[width=1\textwidth]
{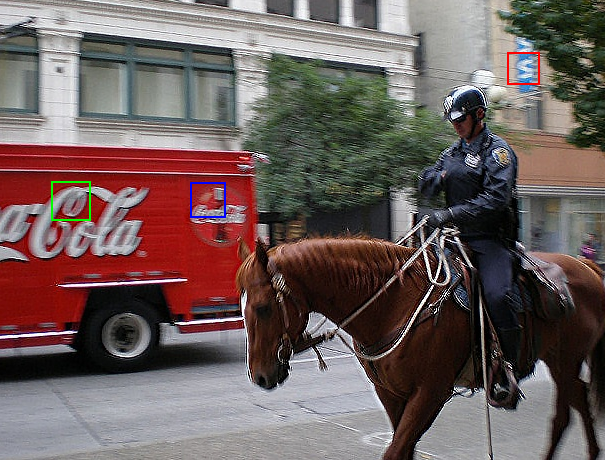}
\put(-2,0.05){ \includegraphics[width=0.23\textwidth]{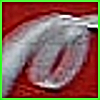}}
\put(21.15,0.05){ \includegraphics[width=0.23\textwidth]{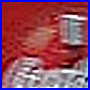}}
\put(44.35,0.05){ \includegraphics[width=0.23\textwidth]{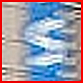}}
\end{overpic}}
\end{minipage}
}
\hspace{-0.3cm}
\subfigure[Sun \etal \cite{sun2015CNN}]{
\begin{minipage}[b]{.23\textwidth}
\centerline{
\begin{overpic}[width=1\textwidth]
{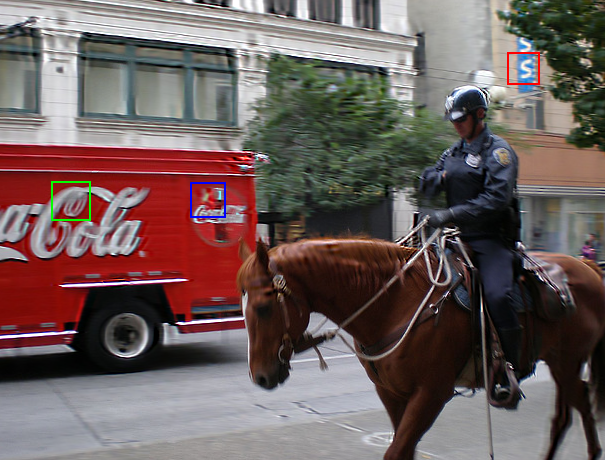}
\put(-2,0.05){ \includegraphics[width=0.23\textwidth]{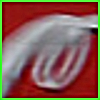}}
\put(21.15,0.05){ \includegraphics[width=0.23\textwidth]{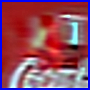}}
\put(44.35,0.05){ \includegraphics[width=0.23\textwidth]{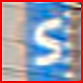}}
\put(67.7,0.05){ \includegraphics[width=0.3\textwidth]{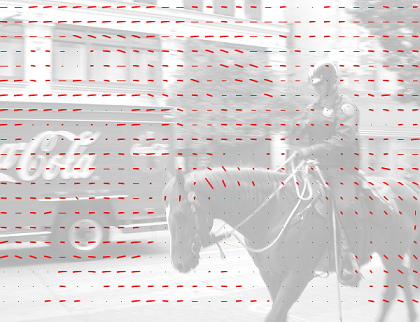}}
\end{overpic}}
\end{minipage}
}
\hspace{-0.3cm}
\subfigure[Ours]{
\begin{minipage}[b]{.23\textwidth}
\centerline{
\begin{overpic}[width=1\textwidth]
{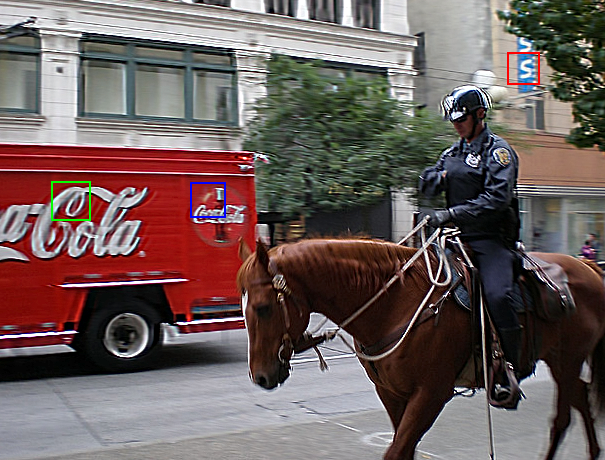}
\put(-2,0.05){ \includegraphics[width=0.23\textwidth]{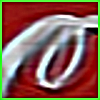}}
\put(21.15,0.05){ \includegraphics[width=0.23\textwidth]{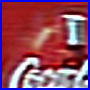}}
\put(44.35,0.05){ \includegraphics[width=0.23\textwidth]{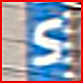}}
\put(67.7,0.05){ \includegraphics[width=0.3\textwidth]{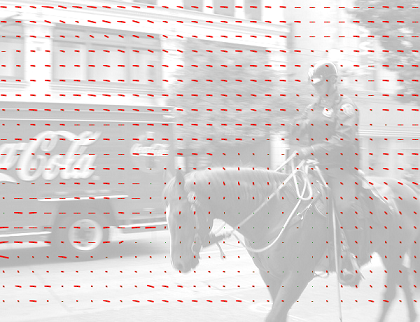}}
\end{overpic}}
\end{minipage}
}
\caption{A blurry image with heterogeneous motion blur from a widely used dataset Microsoft COCO \cite{lin2014mscoco}. Estimated motion flows are shown in the bottom right corner of each image.}
\label{fig:coca}
\vspace{-0.5cm}
\end{figure}

\par
Motion blur in real images has a variety of causes, including camera
\cite{whyte2012non,zheng2013forward} and object motion \cite{hyun2013dynamic,pan2016soft}, leading to blur patterns with complex variations (See Figure \ref{fig:coca} (a)).
In practice, uniform deblurring methods \cite{fergus2006removing,cho2009fast,xu2010twophase} usually fail to remove the non-uniform blur (See Figure \ref{fig:coca} (b)). Most existing non-uniform deblurring methods rely on a specific motion model, such as 3D camera motion modeling \cite{gupta2010single,whyte2012non} and segment-wise motion \cite{levin2006motion,pan2016soft}.
Although a recent method \cite{kim2014segfree} uses a flexible \emph{motion flow} map to handle heterogeneous motion blur, it requires a time-consuming iterative estimator.
In addition to the assumptions about the cause of blur, most existing deblurring methods also rely on predefined priors or manually designed image features.
Most conventional methods \cite{fergus2006removing,levin2011efficient,xu2013unnatural} need to iteratively update the intermediate image and the blur kernel with using these predefined image priors to reduce the ill-posedness.
However, solving these non-convex problems is non-trivial, and many real images do not conform to the assumptions behind a particular model.
Recently, learning-based discriminative methods \cite{chakrabarti2010analyzing,couzinie2013learning} have been proposed to learn blur image patterns and avoid the heavy computational cost of blur estimation. However, their representation and prediction abilities are limited by their manually designed features and simple mapping functions. Although a deep learning based method \cite{sun2015CNN} aimed to overcome these problems, it restrictively conducts the learning process at the patch-level and thus cannot take full advantage of the context information from larger image regions.

\par
In summary, there are three main problems with existing approaches: 1) the range of applicable motion types is limited, 2) manually defined priors and image features may not reflect the nature of the data and 3) complicated and time-consuming optimization and/or post-processing is required. Generally, these problems limit the practical applicability of blur removal methods to real images, as they tend to cause worse artifacts than they cure.

\par
To handle general heterogeneous motion blur, based on the motion flow model, we propose
a deep neural network based method able to directly estimate a
pixel-wise motion flow map from a single blurred image by learning from tens of thousands of examples.
To summarize, the main contributions of this paper are:
\begin{itemize}[itemsep=-1.5pt,topsep=-1.5pt]
\item
We propose an approach to estimate and remove pixel-wise heterogeneous motion blur by training on simulated examples.
Our method uses a flexible blur model and makes almost no assumptions about the underlying images, resulting in effectiveness on diverse data.
\item
We %
introduce a universal FCN for end-to-end estimation of dense heterogeneous motion flow from a single blurry image. Beyond the previous patch-level learning \cite{sun2015CNN}, we directly perform training and testing on the whole image, which utilizes the spatial context over a wider area and estimates a dense motion flow map accurately. Moreover, our method does not require any post-processing.
\end{itemize}

\begin{figure*}[htp]
\centering
\begin{minipage}[b]{.45\textwidth}
\centerline{
\begin{overpic}[trim=1 1 1 1, clip, width=1\textwidth]
{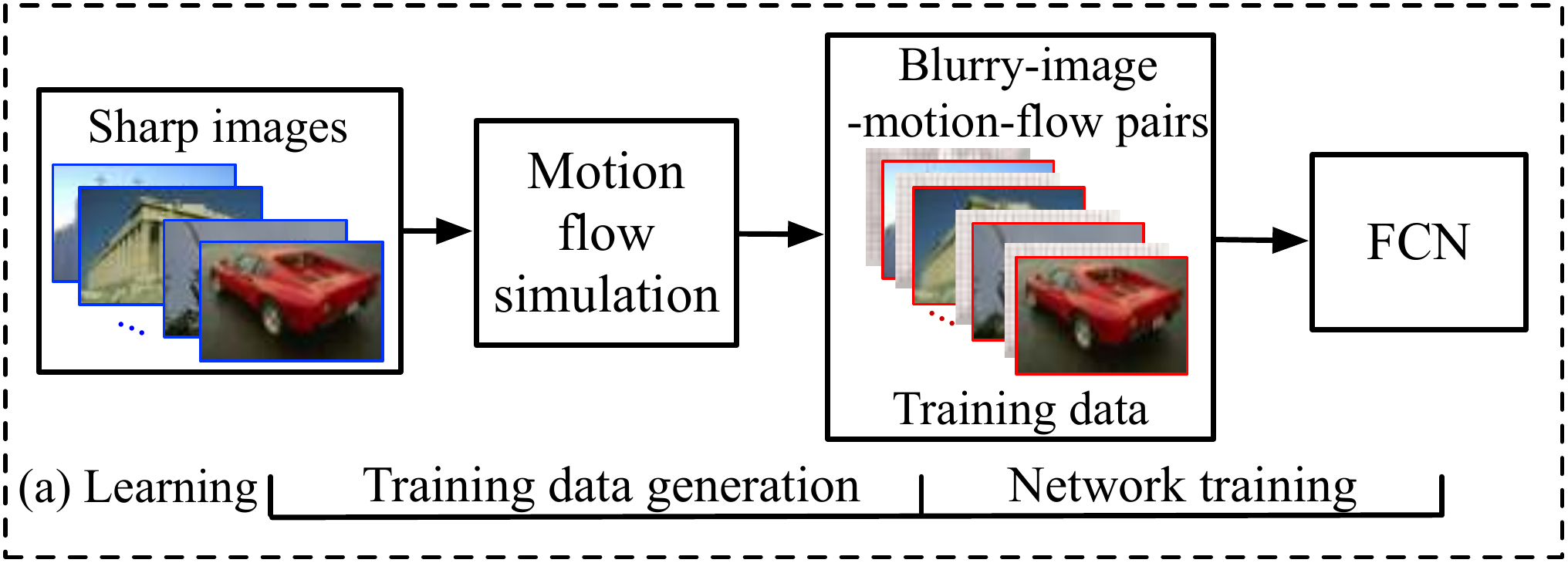}
\end{overpic}}
\end{minipage}
\begin{minipage}[b]{.53\textwidth}
\centerline{
\begin{overpic}[trim=1 1 1 1, clip, width=1\textwidth]
{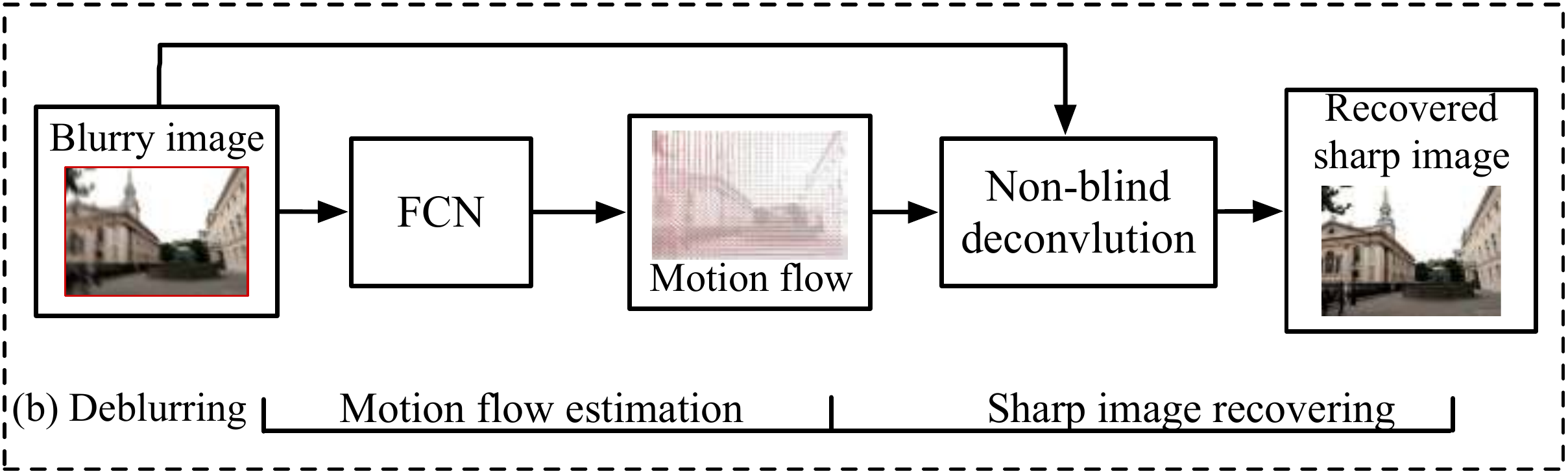}
\end{overpic}}
\end{minipage}
\caption{Overview of our scheme for heterogeneous motion blur removal. (a) We train an FCN using examples based on simulated motion flow maps.
(b) Given a blurry image, we perform end-to-end motion flow estimation using the trained FCN, and then recover the sharp image via non-blind deconvolution.}
\label{fig:overview}
\vspace{-0.2cm}
\end{figure*}

\section{Related Work}
\noindent \textbf{Conventional blind image deblurring}
To constrain the solution space for blind deblurring, a common assumption is that image blur is spatially uniform \cite{chan1998total,cho2009fast,fergus2006removing,levin2011efficient,pan2016dark,gong2016active}.
Meanwhile, numerous image priors or regularizers have been studied to overcome the ill-posed nature of the problem,  such as the total variational regularizer \cite{chan1998total,perrone2014TV}, Gaussian scale mixture priors \cite{fergus2006removing} and $\ell_1/\ell_2$-norms \cite{krishnan2011blind}, $\ell_0$-norms \cite{xu2013unnatural,pan2014text}, and dark channel \cite{pan2016dark} based regularizers.
Moreover, various estimators have been proposed for more robust kernel estimation, such as edge-extraction-based maximum-a-posteriori (MAP) \cite{cho2009fast,sun2013edge}, gradient activation based MAP \cite{gong2016active}, variational Bayesian methods \cite{levin2009understanding,levin2011efficient,zhang2013multi}, \etc. Although these powerful priors and estimators work well on many benchmark datasets, they are often characterised by restrictive assumptions that limit their practical applicability.
\par
\noindent \textbf{Spatially-varying blur removal} To handle  spatially-varying blur, more flexible blur models are proposed. In \cite{tai2011PMP}, a projective motion path model formulates a blurry image as the weighted sum of a set of transformed sharp images, an approach which is which is simplified and extended in \cite{whyte2012non} and \cite{zhang2013non}.
Gupta \etal \cite{gupta2010single} model the camera motion as a motion density function for non-uniform deblurring. Several locally uniform overlapping-patch-based models \cite{hirsch2010EFF,hirsch2011fast} are proposed to reduce the computational burden. Zheng \etal \cite{zheng2013forward} specifically modelled the blur caused by forward camera motion.
To handle blur caused by object motion, some methods \cite{levin2006motion,dai2009removing,hyun2013dynamic,pan2016soft} segment images into areas with different types of blur, and are thus heavily dependent on an accruate segmentation of a blurred image. Recently, a pixel-wise linear motion model \cite{kim2014segfree} is proposed to handle heterogeneous motion blur. Although the motion is assumed to be locally linear, there is no assumption on the latent motion, making it flexible enough to handle an extensive range of possible motion.
\par
\noindent \textbf{Learning based motion blur removing} Recently, learning based methods have been used to achieve more flexible and efficient blur removal.
Some discriminative methods are proposed for non-blind deconvolution based on Gaussian CRF \cite{schmidt2013discriminative}, multi-layer perceptron (MLP) \cite{schuler2013machine}, and deep convolution neural network (CNN) \cite{xu2014deep}, etc, which all require the known blur kernels.
Some end-to-end methods \cite{kim2015accurate,mao2016deconv} are proposed to reconstruct blur-free images, however, they can only handle mild Gaussian blur.
Recently, Wieschollek \etal \cite{WieSchLenHir16Burst} introduce an MLP based blind deblurring method by using information in multiple images with small variations. Chakrabarti \cite{chakrabarti2016neural} trains a patch-based neural network to estimate the frequency information for uniform motion blur removal.
The most relevant work is a method based on CNN and patch-level blur type classification \cite{sun2015CNN}, which also focuses on estimating the motion flow from single blurry image. The authors train a CNN on small patch examples with \emph{uniform} motion blur, where each patch is assigned a single motion label, violating the real data nature and ignoring the correspondence in larger areas.
Many post-processing such as MRF are required for the final dense motion flow.

\section{Estimating Motion Flow for Blur Removal}
\subsection{A Heterogeneous Motion Blur Model}
Letting $*$ denote a general convolution operator, a $P\times Q$ blurred image $\B$ can be modeled as
\begin{equation}
   \B = \K * \I + \N,
   \label{eq:blur_model_2D}
\end{equation}
where $\I$ denotes the latent sharp image, $\N$ refers to additive noise, and $\K$ denotes a heterogeneous motion blur kernel map with different blur kernels for each pixel in $\I$.
Let $\K_{(i,j)}$ represent the
kernel
from $\K$ that operates on a region of the image centered at pixel $(i,j)$.  Thus, at each pixel of $\B$, we have
\begin{equation}
   \B(i,j) = \sum_{i',j'}\K_{(i,j)}(i',j')\I(i+i', j+j').
   \label{eq:blur_model_2D_pixel}
\end{equation}
If we define an operator $\vec(\cdot)$ which vectorises a matrix and let
$\by = \vec(\B)$, $\bx = \vec(\I)$ and $\bn = \vec(\N)$ then
\eqref{eq:blur_model_2D} can also be represented as
\begin{equation}
\by=\bH(\K)\bx + \bn,
\label{eq:blur_model_mv}
\end{equation}
where $\bH(\K) \in \mbR^{PQ\times PQ}$\footnote{For simplicity, we assume $\I$ and $\B$ have the same size.}and each row corresponds to a blur kernel located at each pixel (\ie $\K_{(i,j)}$).

\subsection{Blur Removal via Motion Flow Estimation}\label{sec:overview}
Given a blurry image $\B$, our goal is to estimate the blur kernel $\K$ and recover a blur-free latent image $\I$ through non-blind deconvolution that can be performed by solving a convex problem (Figure \ref{fig:overview} (b)). As mentioned above, kernel estimation is the most difficult and crucial part.

\begin{figure}[htp]
\centering
\subfigure[Motion blur and motion flow]
{\begin{minipage}[b]{.21\textwidth}
\centerline{
\begin{overpic}[width=1\textwidth]
{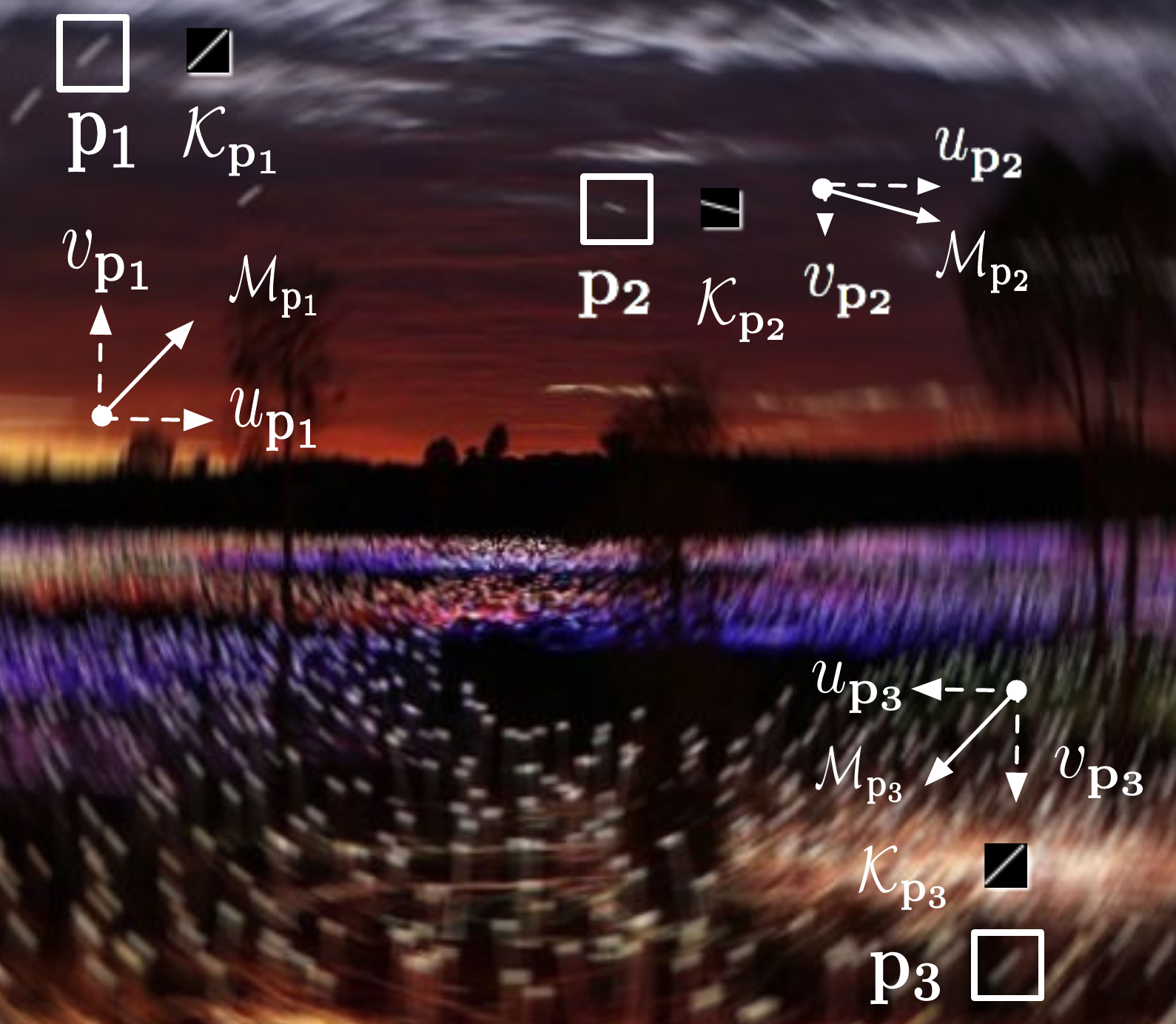}
\put(-1.5,0.5){ \includegraphics[width=0.4\textwidth]{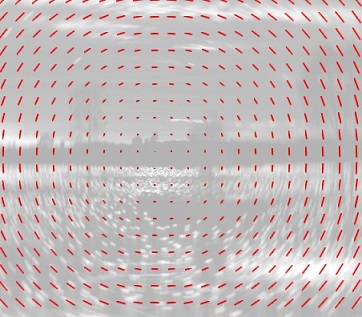}}
\end{overpic}}
\end{minipage}
}
\hspace{0.5cm}
\subfigure[Domain of motion]
{\includegraphics[width=0.15\textwidth]{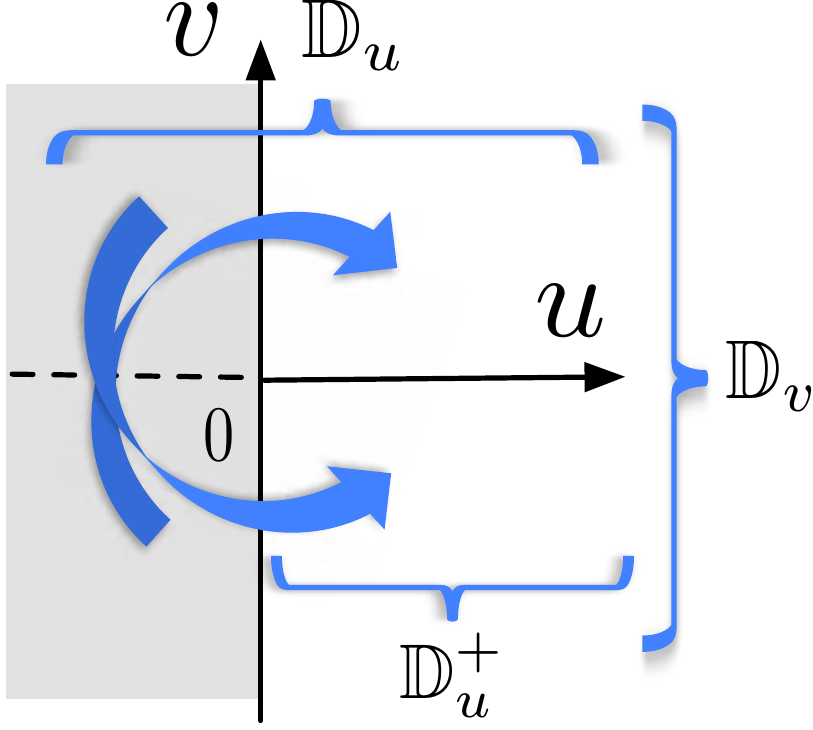}
}
\caption{Motion blur and motion vector. (a) An example with blur cause by clock-wise rotation. Three examples of the blur pattern, linear blur kernel and motion vector are shown. The blur kernels on $\bp_1$ and $\bp_3$ caused by motions with opposite directions and have the same appearance. (b) Illustrations of the feasible domain of motion flow.}
\label{fig:motion2blur}
\vspace{-0.1cm}
\end{figure}

\begin{figure*}[t!]
\centering
\includegraphics[width=1\textwidth]{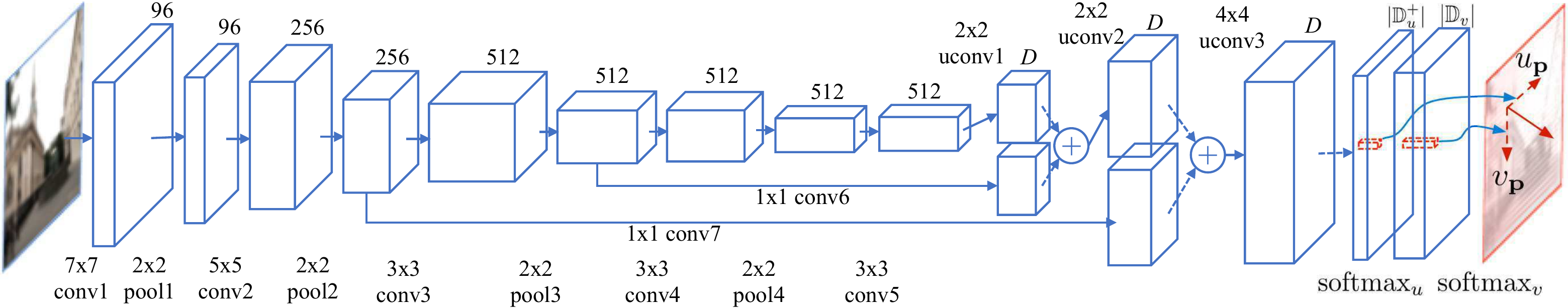}
\caption{Our network structure. A blurred image goes through layers and produces a pixel-wise dense motion flow map. {conv} means a convolutional layer and {uconv} means a fractionally-strided convolutional (deconvolutional) layer, where $n\times n$ for each {uconv} layer denotes that the up-sampling size is $n$. Skip connections on top of {pool2} and {pool3} are used to combine features with different resolutions.}
\label{fig:net}
\vspace{-0.3cm}
\end{figure*}

\par
Based on the model in \eqref{eq:blur_model_2D} and \eqref{eq:blur_model_2D_pixel}, heterogeneous motion blur can be modeled by a set of blur kernels, one associated with each pixel and its motion.
By using a linear motion model to indicate each pixel's motion during imaging process \cite{kim2014segfree}, and letting $\bp=(i,j)$ denote a pixel location, the motion at pixel $\bp$, can be represented by a 2-dimensional \emph{motion vector} $(u_\bp, v_\bp)$, where  $u_\bp$ and $v_\bp$ represent the movement in the horizontal and vertical directions, respectively (See Figure \ref{fig:motion2blur} (a)).
By a slight abuse of notation we express this as $\bm_\bp=(u_\bp, v_\bp)$, which characterizes the movement at pixel  $\bp$ over the exposure time. If we have the feasible domain $u_\bp\in \mbD_u$ and $v_\bp\in \mbD_v$, then $\bm_{\bp}\in \mbD_u \times \mbD_v$, but will be introduced in detail later.
As shown in Figure \ref{fig:motion2blur}, the blur kernel on each pixel appears as a line trace with nonzero components only along the motion trace.
As a result, the motion blur $\K_{\bp}$ in \eqref{eq:blur_model_2D_pixel} can be expressed as \cite{brusius2011blind}:
\begin{equation}
 \K_{\bp}(i',j') = \left\{
      \begin{array}{l}
        \!\! ~~~~~~~0, ~~~~~~~\text{if}~ \|(i',j')\|_2\geq\frac{\|\M_\bp\|_2}{2},\\
        \!\! \frac{1}{\|\M_\bp\|_2}\delta(v_\bp i'\!-\!u_\bp j'), ~~~\text{otherwise},\\
      \end{array}
      \right.
      \label{eq:pix_ker}
\end{equation}
where $\delta(\cdot)$ denotes the Dirac delta function.
We thus can achieve heterogeneous motion blur estimation by estimating the motion vectors on all pixels, the result of which is $\M$, which is referred as \emph{motion flow}. For convenience of expression,
we let $\M\!=\!(\U,\V)$, where $\U$ and $\V$ denote the motion maps in the horizontal and vertical directions, respectively. For any pixel $\bp=(i,j)$, we define
$\bm_\bp=(\U(i,j), \V(i,j))$ with $\U(i,j) = u_\bp$ and $\V(i,j)= v_\bp$.

As shown in Figure \ref{fig:overview} (b), given a blurred image and the estimated motion flow, we can recover the sharp image by solving an non-blind deconvolution problem \[\min_{\bx}\|\by-\bH(\K)\bx\|_2^2+\Omega(\bx)\] with regularizer $\Omega(\bx)$ on the unknown sharp image. In practice, we use a Gaussian mixture model based regularizer as $\Omega(\bx)$
\cite{zoran2011EPLL,sun2015CNN}.

\subsection{Learning for Motion Flow Estimation}
The key contribution of our work is to show how to obtain the motion flow field that results in the pixel-wise motion blur.  To do so we train a FCN to directly estimate the motion flow field from the blurry image.

\par
Let $\{(\B^t, \M^t)\}_{t=1}^T$ be a set of blurred-image and motion-flow-map pairs, which we take as our training set.
Our task is to learn an end-to-end mapping function $\M=f(\B)$ from any observed blurry image $\B$ to the underlying motion flow $\M$.
In practice, the challenge is that obtaining the training ground-truth dense motion flow for
sufficiently many and varied real blurry images is infeasible. Human labeling is impossible, and training from automated methods for image deblurring would defeat the purpose.
To overcome this problem, we generate the training set by simulating motion flows maps.
(See section \ref{sec:pseudo-phyEngine}). Specifically, we collect a set of sharp images $\{\I^n\}$, simulate $T$ motion flows $\{\M^t\}$ in total for all images in $\{\I^n\}$, and then generate $T$
 blurred images $\{\B^t\}$ based on the models in \eqref{eq:blur_model_2D} and \eqref{eq:pix_ker} (See Figure \ref{fig:overview} (a)).
\par
\noindent \textbf{Feasible domain of motion flow}
To simplify the training process, we train the FCN over a discrete output domain.
Interestingly, classification on discrete output space has achieved some
impressive results for some similar applications, \eg optical flow estimation \cite{walker2015optical} and surface normal prediction \cite{wang2015surface}. In our work,
we adopt an integer domain for both $\U$ and $\V$,
and treat the mapping $\M=f(\B)$ as a multi-class classification problem.
Specifically, we uniformly discretize the motion values as integers with a 1 (pixel) interval, which provides a high-precision approximation to the latent continuous space. As a result, by assuming
the maximum movements in the
horizontal and vertical directions to be $u_{max}$ and $v_{max}$, respectively, we have $\mbD_u=\{u | u\in\mbZ, |u|\leq u_{max}\}$ and $\mbD_v=\{v | v\in\mbZ, |v|\leq v_{max}\}$, where $\mbZ$ denotes the integer domain.
\par
As shown in Figure \ref{fig:motion2blur} (a), any linear blur kernel is symmetric. Any two motion vectors with same length and opposite directions, \eg $(u_\bp, v_\bp)$ and $(-u_\bp, -v_\bp)$, generate the same blur pattern, which may confuse the learning process. We thus further restrict the motion in the horizontal direction to be nonnegative as shown in Figure \ref{fig:motion2blur} (b), \ie $u_\bp\in \mbD_u^+=\{u | u\in\mbZ_0^+, |u|\leq u_{max}\}$, by letting $(u_\bp, v_\bp) = \phi(u_\bp, v_\bp)$ where
\begin{equation}
   \phi(u_\bp, v_\bp) = \left\{
      \begin{array}{l}
        \!\! (-u_\bp, -v_\bp), ~~~\text{if}~ u_\bp<0,\\
        \!\! ~~~(u_\bp, v_\bp), ~~~~~~\text{otherwise}.\\
      \end{array}
      \right.
\end{equation}

\begin{figure*}[!t]
\centering
\subfigure[Sharp Image]{
\begin{minipage}[b]{.12\textwidth}
\centerline{
\begin{overpic}[width=1\textwidth]
{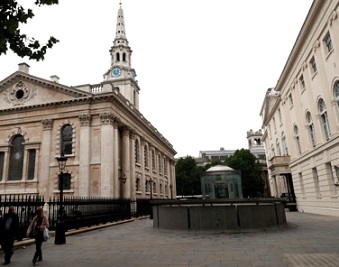}
\put(-0,80){ \includegraphics[width=0.5\textwidth]{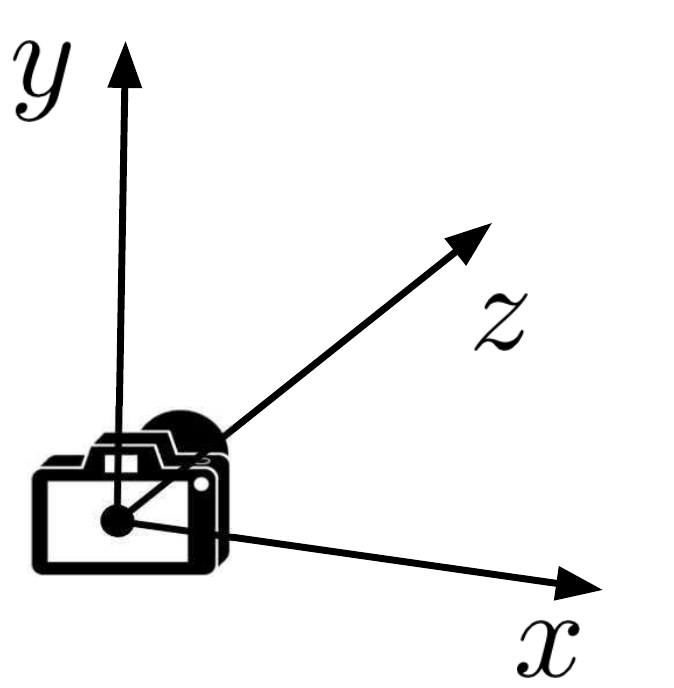}}
\put(-4,0.3){ \includegraphics[width=0.25\textwidth]{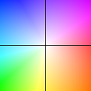}}
\end{overpic}}
\end{minipage}
}
\hspace{-0.3cm}
\subfigure[$x$ and $y$-axis translation]{
\begin{minipage}[b]{.21\textwidth}
\centerline{
\begin{overpic}[width=1\textwidth]
{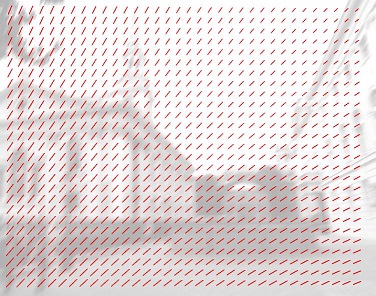}
\put(-2,47){ \includegraphics[width=0.3\textwidth]{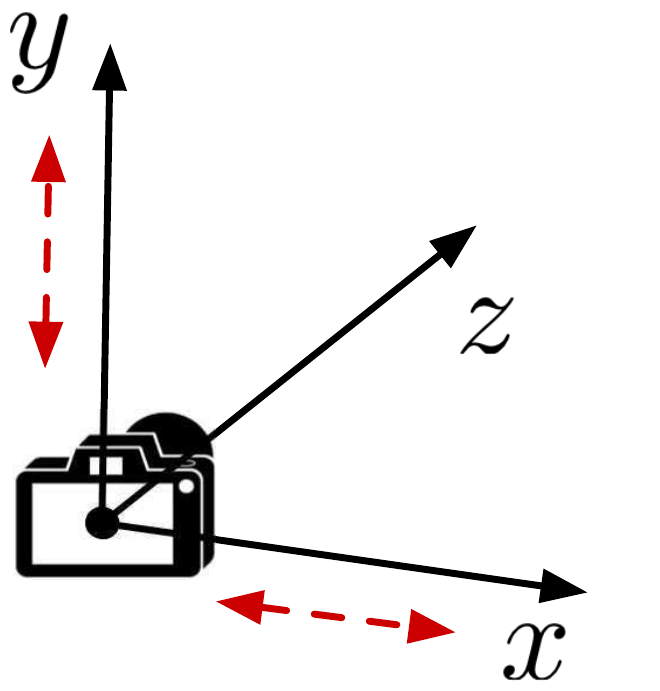}}
\put(-2,0.5){ \includegraphics[width=0.27\textwidth]{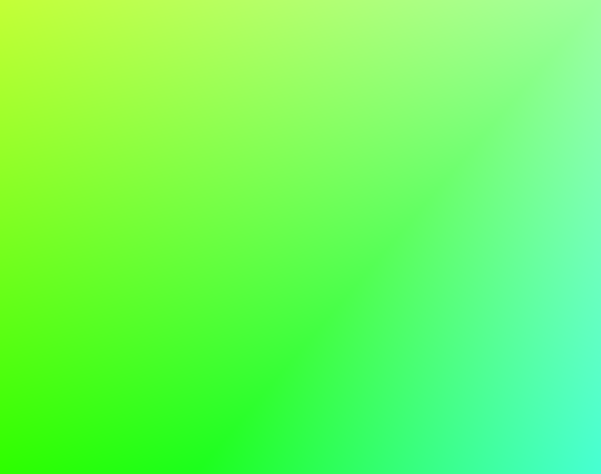}}
\put(47,0.5){ \includegraphics[width=0.5\textwidth]{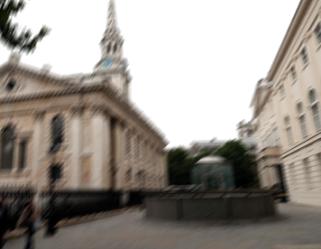}}
\end{overpic}}
\end{minipage}
}
\hspace{-0.3cm}
\subfigure[$z$-axis translation]{
\begin{minipage}[b]{.21\textwidth}
\centerline{
\begin{overpic}[width=1\textwidth]
{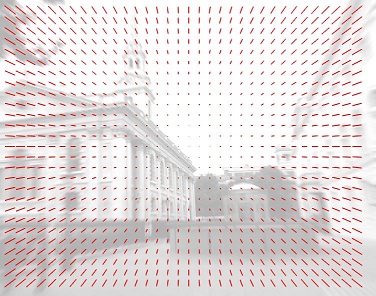}
\put(-2,47){ \includegraphics[width=0.3\textwidth]{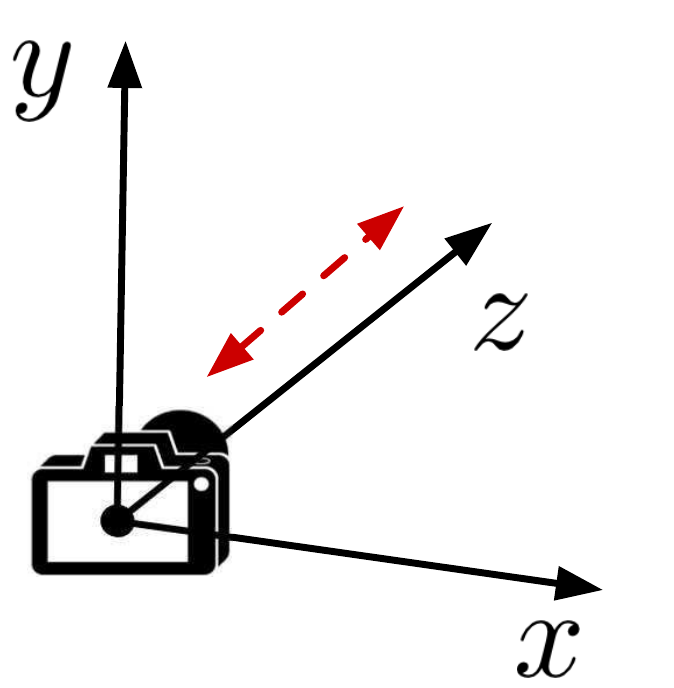}}
\put(-2,0.5){ \includegraphics[width=0.27\textwidth]{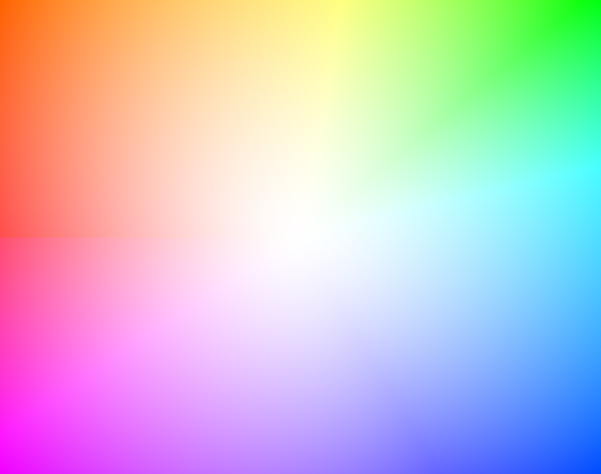}}
\put(47,0.5){ \includegraphics[width=0.5\textwidth]{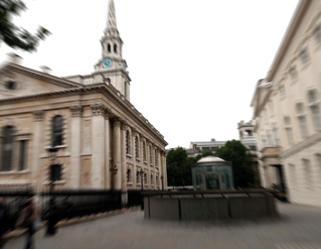}}
\end{overpic}}
\end{minipage}
}
\hspace{-0.3cm}
\subfigure[$z$-axis rotation]{
\begin{minipage}[b]{.21\textwidth}
\centerline{
\begin{overpic}[width=1\textwidth]
{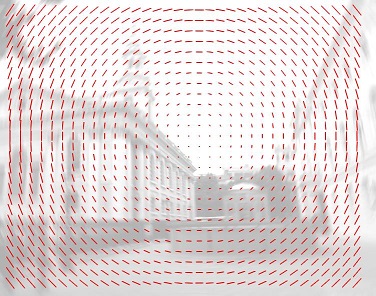}
\put(-2,47){ \includegraphics[width=0.3\textwidth]{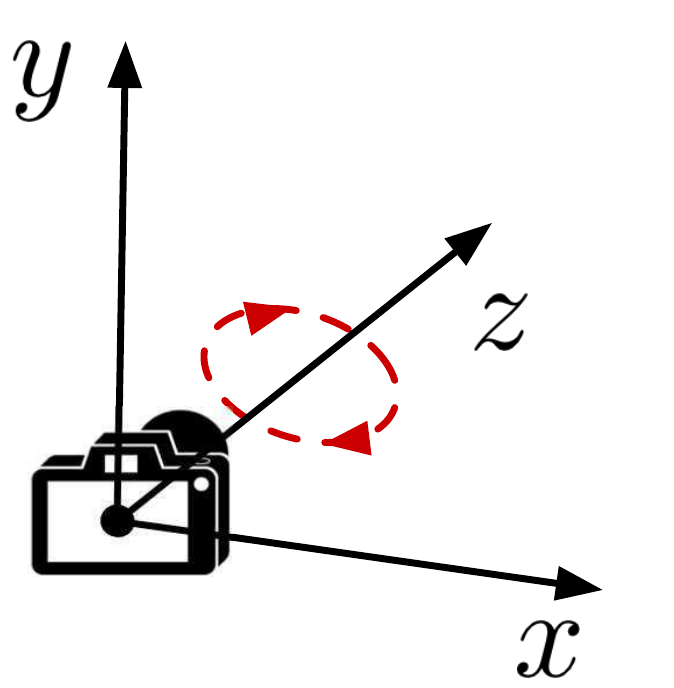}}
\put(-2,0.5){ \includegraphics[width=0.27\textwidth]{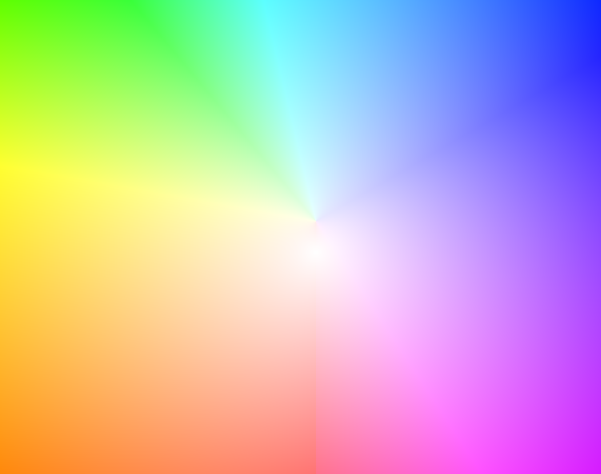}}
\put(47,0.5){ \includegraphics[width=0.5\textwidth]{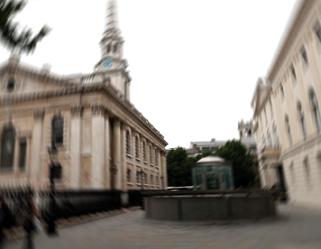}}
\end{overpic}}
\end{minipage}
}
\hspace{-0.3cm}
\subfigure[Arbitrary sampled motion]{
\begin{minipage}[b]{.21\textwidth}
\centerline{
\begin{overpic}[width=1\textwidth]
{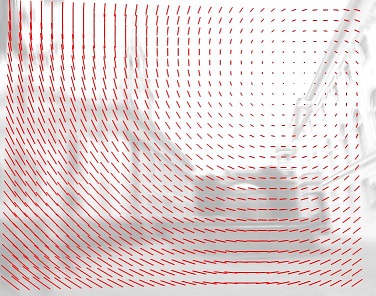}
\put(-2,47){ \includegraphics[width=0.3\textwidth]{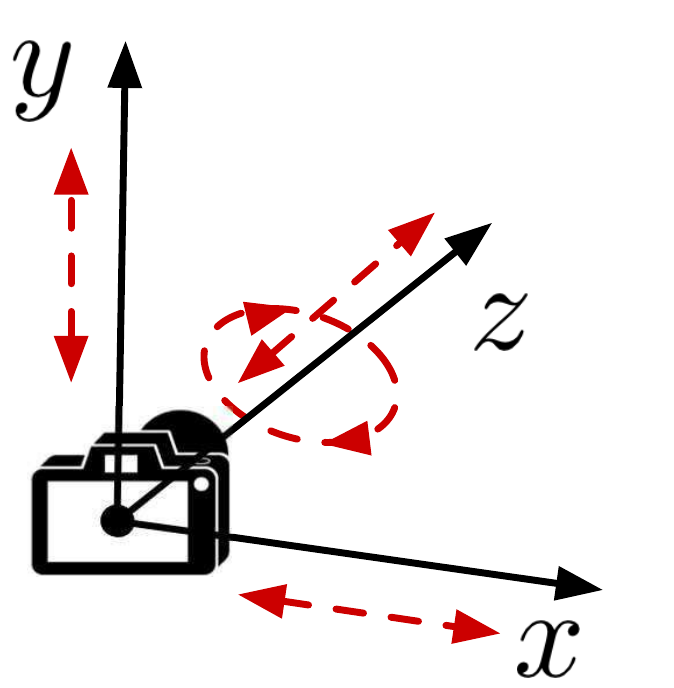}}
\put(-2,0.5){ \includegraphics[width=0.27\textwidth]{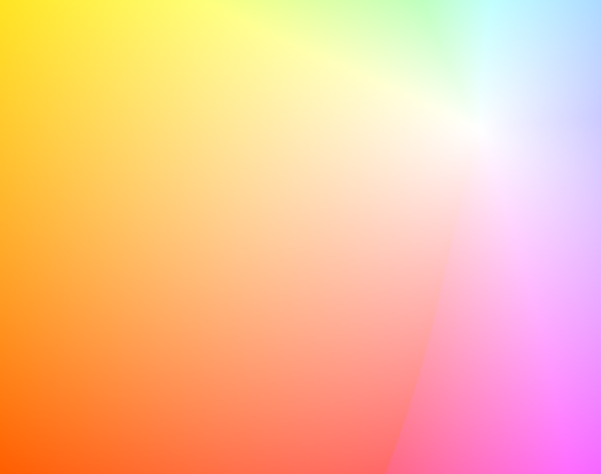}}
\put(47,0.5){ \includegraphics[width=0.5\textwidth]{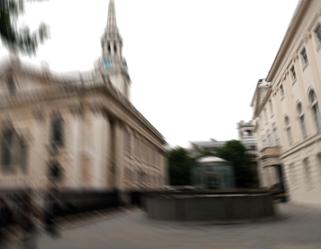}}
\end{overpic}}
\end{minipage}
}
\caption{Demonstration of the motion flow simulation. (a) A sharp example image and the coordinate system of camera. (b)-(c) The sampled motion flow and the corresponding blurred image by simulating the translation along $x$ and $y$-axes ($\M_{T_x}+\M_{T_y}$), translation along $z$-axis ($\M_{T_z}$) and rotation around $z$-axis ($\M_{R_z}$), respectively. (d) A sample based on the model considering all components in \eqref{eq:mf_decomp}.}
\label{fig:phyeng}
\vspace{-0.2cm}
\end{figure*}

\section{Dense Motion Flow Estimation}

\subsection{Network Design}
The goal of this FCN network is to achieve the end-to-end mapping from a blurry image to its corresponding motion flow map. Given any RGB image with the arbitrary size $P\times Q$, the FCN is used to estimate a motion flow map $\M=(\U,\V)$ with the same size to the input image, where $\U(i,j)\in \mbD_u^+$ and $\V(i,j)\in \mbD_v$, $\forall i,j$. For convenience, we let $D=|\mbD_u^+| + |\mbD_v|$ denote the total number of labels for both $\U$ and $\V$.
Our network structure is similar to the FCN in \cite{long2015FCN}. As shown in Figure \ref{fig:net}, we use 7 convolutional ({conv}) layers and 4 max-pooling ({pool}) layers as well as $3$ {uconv} layers to up-sample the prediction maps.
Following \cite{wang2016generative}, {uconv} denotes the fractionally-strided convolution, a.k.a. deconvolution.
We use a small stride of 1 pixel for all convolutional layers.
The {uconv} layers are initialized with bilinear interpolation and used to up-sample the activations. We also add skip connections which combine the information from different layers as shown in Figure \ref{fig:net}.

\par
The feature map of the last uconv layer (conv7 + uconv2) is a $P \times Q \times D$ tensor with the top $|\mbD_u^+|$ slices of feature maps ($P \times Q \times |\mbD_u^+|$) corresponding to the estimation of $\U$, and the remaining $|\mbD_v|$ slices of feature maps ($P \times Q \times |\mbD_v|$) corresponding to the estimation of $\V$. Two separate soft-max layers are applied to those two parts respectively to obtain the posterior probability estimation of both channels. Let $F_{u,i,j}(\B)$ represent the probability that the pixel at $(i,j)$ having a movement $u$ along the horizontal direction, and $F_{v,i,j}(\B)$ represent the probability that the pixel at $(i,j)$ having a movement $v$ along the vertical direction, we then use the sum of the cross entropy loss from both channels as the final loss function:
\begin{eqnarray}
\begin{split}
   L(\B,\M)\! =\! &-\!\sum_{i=1}^P\sum_{j=1}^Q \! \sum_{u\in \mbD_u^+}\! \mb1(\U(i,j)=u) \log(\F_{u,i,j}(\B)) \\
     & -\!\sum_{i=1}^P\sum_{j=1}^Q \!\sum_{v\in \mbD_v} \mb1(\V(i,j)=v) \log(\F_{v,i,j}(\B)),
\end{split}
   \label{eq:loss_fun}
   \nonumber
\end{eqnarray}
where $\mb1$ is an indicator function.

\subsection{Simulate Motion Flow for Data Generation} \label{sec:pseudo-phyEngine}
The gist of this section is generating a dataset that contains realistic blur patterns on diverse images for training. Although an i.i.d. random sampling may generate very diverse training samples, since the realistic motion flow preserves some properties such as piece-wise smoothness, we aim to design a simulation method
to generate motion flows reflecting the natural properties of the movement in imaging process.
Although the object motion \cite{hyun2013dynamic} can lead to heterogeneous motion blur in real images, our method only simulates the motion flow caused by camera motion for learning.
Even so, as shown in Section \ref{sec:exp_real}, data generated by our method can also give the model certain ability to handle object motion.

\par
For simplicity, we generate a 3D coordinate system where the origin at the camera's optical center, the $xy$-plane is aligned with the camera sensors, and the $z$-axis is perpendicular to the $xy$-plane, as shown in Figure \ref{fig:phyeng}.
Since our objective is the motion flow on an image grid, we directly simulate the motion flow projected on 2D image instead of the 3D motion trajectory \cite{whyte2012non}. Considering the ambiguities caused by rotations around $x$ and $y$ axis \cite{gupta2010single},
we simulate a motion flow $\M$ by sampling four additive components:
\begin{equation}
   \M = \M_{T_x}+\M_{T_y}+\M_{T_z}+\M_{R_z},
   \label{eq:mf_decomp}
\end{equation}
where $\M_{T_x}$, $\M_{T_y}$ and $\M_{T_z}$ denote the motion flows associated with the translations along $x$, $y$ and $z$ axis, receptively, and $\M_{R_z}$ represents the motion from the rotation around $z$ axis. We generate each element as the following.

\par
\noindent \textbf{Translation along $x$ or $y$ axis}
We describe the generation of $\M_{T_x}$ as an example.
We first sample a central pixel $\bp_{T_x}=(i_{T_x}, j_{T_x})$ on image plane, a basic motion value $t_{T_x}$ and a acceleration coefficient $r_{T_x}$. Then $\M_{T_x}=(\U_{T_x}, \V_{T_x})$ can be generated as the following $\U_{T_x}(i,j) = (i-i_{T_x})r_{T_x} + t_{T_x}, \V_{T_x}(i,j) = 0$.
$\M_{T_y}$ can be generated in a similar way.

\noindent \textbf{Translation along $z$ axis} %
The translation along $z$ axis usually causes radial motion blur pattern towards the vanishing point \cite{zheng2013forward}.
By ignoring the semantic context and assuming a simple radial pattern, $\M_{T_z}$ can be generated by $\U_{T_z}(i,j) = t_{T_z} d(i,j)^\zeta (i-i_{T_z}), \V_{T_z}(i,j) = t_{T_z} d(i,j)^\zeta (j-j_{T_z})$
where $\bp_{T_z}$ denotes a sampled vanishing point, $d(i,j) = \|(i,j)-\bp_{T_z}\|_2$ is the distance from any pixel $(i,j)$ to the vanishing point,
$\zeta$ and $t_{T_z}$ are used to control the shape of radial patterns, which reflects the moving speed.

\noindent \textbf{Rotation around $z$ axis}
We first sample a rotation center $\bp_{R_z}$ and an angular velocity $\omega$, where $\omega>0$ denotes the clockwise rotation. Let $d(i,j) = \|(i,j)-\bp_{R_z}\|_2$. The motion magnitude at each pixel is $s(i,j)=2d(i,j)\tan(\omega/2)$. By letting $\theta(i,j)=\text{atan}[(i-i_{R_z})/(j-j_{R_z})] \in [-\pi, \pi]$, motion vector at pixel $(i,j)$ can be generated as $\U_{R_z}(i,j) = s(i,j) \cos(\theta(i,j)-\pi/2), \V_{R_z}(i,j) = s(i,j) \sin(\theta(i,j)-\pi/2)$.

\par
We place uniform priors over all the parameters corresponding to the motion flow simulation
as $\text{Uniform}(\alpha, \beta)$. More details can be found in supplementary materials. Note that the four components in \eqref{eq:mf_decomp} are simulated in continuous domain and are then discretized as integers.

\par
\noindent \textbf{Training dataset generation}
We use 200 training images with sizes around $300\times 460$ from the dataset BSD500 \cite{bsd500} as our sharp image set $\{\I^n\}$. We then independently simulate 10,000 motion flow maps $\{\M^t\}$ with ranges $u_{max}=v_{max}=36$ and assign each $\I^n$ 50 motion flow maps without duplication.
The non-blurred images $\{\I^n\}$ with $\U(i,j)=0$ and $\V(i,j)=0$, $\forall i,j$ are used for training.
As a result we have a dataset with 10,200 blurred-image-motion-flow pairs $\{\B^t, \M^t\}$ for training.

\begin{table*}[!t]
\centering
\small
\caption{Evaluation on motion blur estimation. Comparison on PSNR and SSIM of the recovered images with the estimated blur kernel. The best results are bold-faced.}
\vspace{0.1cm}
\label{tab:k_eval}
\small
\begin{tabular}{c|c|c|c c c c c c}
\hline
  Dataset   & Metric  &  GT $\K$  & Xu and Jia \cite{xu2010twophase} & Whyte \etal \cite{whyte2012non} & Xu \etal \cite{xu2013unnatural} &  noMRF \cite{sun2015CNN}& patchCNN \cite{sun2015CNN} & Ours \\ \hline
BSD-S & PSNR  & 23.022 & 17.773  & 17.360 & 18.351 &  20.483& 20.534 &\bf 21.947 \\
      & SSIM  & 0.6609 & 0.4431 & 0.3910 & 0.4766 &  0.5272& 0.5296 &\bf 0.6309 \\ \hline
BSD-M & PSNR  & 24.655 & 19.673 & 18.451  & 20.057 &  22.789  &22.9683& \bf 23.978 \\
      & SSIM  & 0.7481 & 0.5661 & 0.5010 & 0.5973 & 0.6666 & 0.6735 & \bf 0.7249 \\ \hline
\end{tabular}
\end{table*}

\section{Experiments}
We implement our model based on Caffe \cite{jia2014caffe} and train it by stochastic gradient descent with momentum and batch size 1. In the training on the dataset simulated on BSD, we use a learning rate of $10^{-9}$ and a step size of $2\times 10^5$. The training converges after 65 epochs.

\subsection{Datasets and Evaluation Metrics}
\noindent \textbf{Datasets}
We conduct the experiments on both \emph{synthetic datasets} and \emph{real-world images}.
Since ground truth motion flow and sharp image for real blurry image are difficult to obtain, to perform general quantitative evaluation, we first generate two synthetic datasets, which both contain 300 blurred images, with 100 sharp images randomly picked from BSD500 \cite{bsd500}\footnote{No overlapping with the training dataset.}, and 3 different motion flow maps for each. Note that no two motion flow maps are the same. We simulate the motion flow with $u_{max}=v_{max}=36$, which is same as in the training set. For fairness to the method \cite{sun2015CNN} with a smaller output space, we also generate relative mild motion flows for the second dataset with $u_{max}=v_{max}=17$. These two are referred as \textbf{BSD-S} and \textbf{BSD-M}, respectively. In addition, we evaluate the generalization ability of the proposed method using two synthetic datasets (\textbf{MC-S} and \textbf{MC-M}) with 60 blurry images generated from 20 sharp images from Microsoft COCO \cite{lin2014mscoco} and above motion flow generation setting.

\par
\noindent \textbf{Evaluation Metrics} For evaluating the accuracy of estimated motion flow, we measure the mean-squared-error (\textbf{MSE}) of the motion flow map.
Specifically, given an estimated motion flow $\widehat{\M}$ and the ground truth $\M$, the MSE is defined as $\frac{1}{2|M|}\sum_{i,j} ( (\U(i,j)-\widehat{\U}(i,j))^2 + ((\V(i,j)-\widehat{\V}(i,j))^2$, where $|\M|$ denotes the number of motion vectors in $\M$.
For evaluation of the image quality, we adopt peak signal-to-noise-ratio (\textbf{PSNR}) and structural similarity index (\textbf{SSIM}) \cite{wang2004SSIM}.

\begin{figure}[!t]
\centering
\begin{minipage}[b]{.12\textwidth}
\centerline{
\begin{overpic}[width=1\textwidth]
{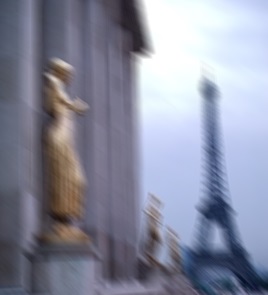}
\put(2,3){\scriptsize\color{white}{\bf (a) Blurry image}}
\end{overpic}}
\end{minipage}
\hspace{-0.2cm}
\begin{minipage}[b]{.12\textwidth}
\centerline{
\begin{overpic}[width=1\textwidth]
{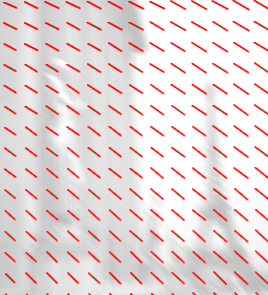}
\put(2,3){\scriptsize\color{black}{\bf (b) Ground truth}}
\end{overpic}}
\end{minipage}
\hspace{-0.2cm}
\begin{minipage}[b]{.12\textwidth}
\centerline{
\begin{overpic}[width=1\textwidth]
{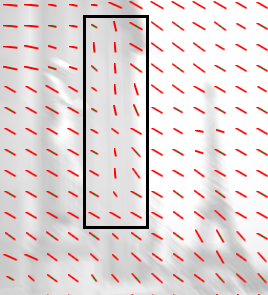}
\put(-1,3){\scriptsize\color{black}{\bf (c) \cite{sun2015CNN}, MSE:16.68 }}
\end{overpic}}
\end{minipage}
\hspace{-0.2cm}
\begin{minipage}[b]{.12\textwidth}
\centerline{
\begin{overpic}[width=1\textwidth]
{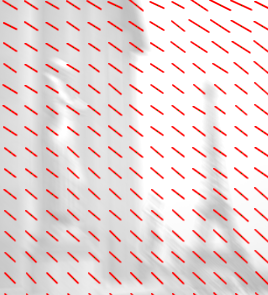}
\put(2,3){\scriptsize\color{black}{\bf (d) Ours, MSE:1.05 }}
\end{overpic}}
\end{minipage}
\caption{A motion flow estimation example on a synthetic image in BSD-M. The method of Sun \etal \cite{sun2015CNN} is more sensitive to the image content (See the black box in (c)).}
\label{fig:syn_mf}
\end{figure}

\subsection{Evaluation of Motion Flow Estimation}

We first compare with the method of Sun \etal (``patchCNN'') \cite{sun2015CNN}, which is the only method with available code for estimating motion flow from blurry images\footnote{The code of the other motion flow based method \cite{kim2014segfree} is unavailable.}. This method performs training and testing on small image patches, and uses MRF to improve the accuracy on the entire image. Its version without MRF post-processing (``noMRF'') is also compared, where the soft-max output is directly used to get the motion flow as in our method.
Table \ref{tab:mse-m} shows the average MSE of the estimated motion flow maps on all images in BSD-S and BSD-M.
It is noteworthy that, even without any post-processing such as MRF or CRF, the comparison manifests the high quality of our estimated motion flow maps.
Furthermore, our method can still produce accurate motion flow even on the more challenging BSD-S dataset, on which the accuracies of the patch based method \cite{sun2015CNN} decrease significantly.
We also show an example of the the estimated motion flows in Figure \ref{fig:syn_mf}, which shows that our result preserves a smooth motion flow very similar to the ground truth, and the method of Sun \etal \cite{sun2015CNN} is more sensitive to the image contents.
From this example, we can see that the method of Sun \etal \cite{sun2015CNN} generally underestimates the motion values and produces errors near the strong edges, maybe because its patch-level processing is confused by the strong edges and ignores the blur pattern context in a larger area.
\begin{table}[htp]
\centering
\caption{Evaluation on motion flow estimation (MSE). The best results are bold-faced.}
\vspace{0.05cm}
\label{tab:mse-m}
\small
\begin{tabular}{c|ccc}
\hline
Dataset & patchCNN \cite{sun2015CNN} &  noMRF \cite{sun2015CNN}  & Ours \\ \hline
BSD-S   & 50.1168 &  54.4863 &  \bf 6.6198\\
BSD-M   & 15.6389 &  20.7761 &  \bf 5.2051 \\ \hline
\end{tabular}
\vspace{-0.3cm}
\end{table}

\par
To compare with other blind deblurring methods of Xu and Jia \cite{xu2010twophase}, Xu \etal \cite{xu2013unnatural} and Whyte \etal \cite{whyte2012non}, which do not estimate the motion flow, we directly evaluate the quality of the image recovered using their estimated blur kernel. For fairness, we use the same non-blind deconvolution method with least square loss function and a Gaussian mixture model prior \cite{zoran2011EPLL} to recover the sharp image.
As the non-blind deconvolution method may limit the recovering quality, we evaluate the images recovered using the ground truth motion flow as reference. Table \ref{tab:k_eval} shows the average values on all images in each dataset, which shows that our method produce significantly better results than the others.

\begin{figure*}[!t]
\centering
\begin{minipage}[b]{.24\textwidth}
\centerline{
\begin{overpic}[width=1\textwidth]
{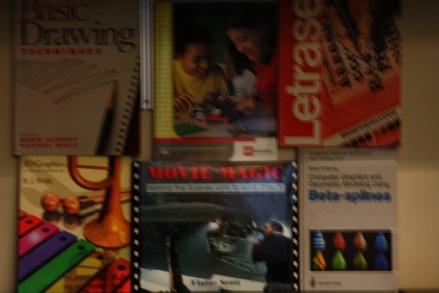}
\put(2,3){\footnotesize\color{white}{\bf (a) Blurry image}}
\end{overpic}}
\end{minipage}
\begin{minipage}[b]{.24\textwidth}
\centerline{
\begin{overpic}[width=1\textwidth]%
{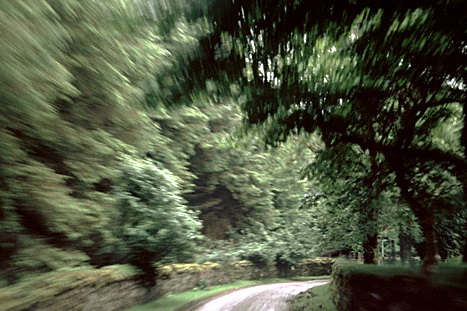}
\put(2,3){\footnotesize\color{white}{\bf (b) Blurry image}}
\end{overpic}}
\end{minipage}
\begin{minipage}[b]{.24\textwidth}
\centerline{
\begin{overpic}[width=1\textwidth]
{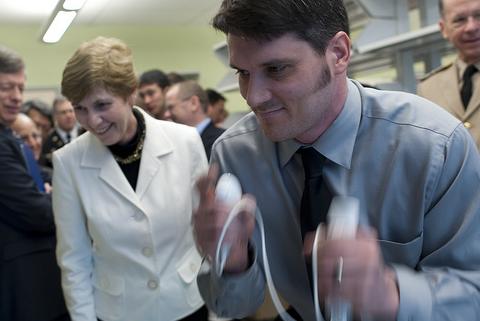}
\put(2,3){\footnotesize\color{white}{\bf (c) Blurry image}}
\end{overpic}}
\end{minipage}
\begin{minipage}[b]{.24\textwidth}
\centerline{
\begin{overpic}[width=1\textwidth]
{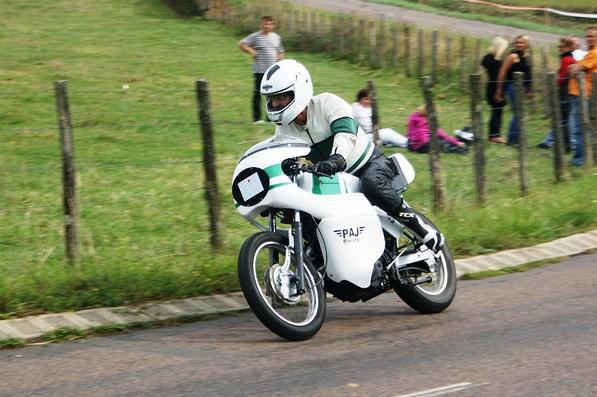}
\put(2,3){\footnotesize\color{white}{\bf (d) Blurry image}}
\end{overpic}}
\end{minipage}
\begin{minipage}[b]{.24\textwidth}
\centerline{
\begin{overpic}[width=1\textwidth]
{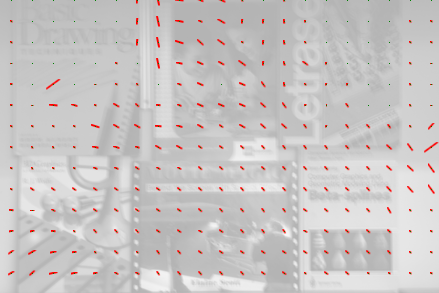}
\put(2,3){\footnotesize\color{black}{\bf (e) Motion flow of \cite{sun2015CNN} }}
\end{overpic}}
\end{minipage}
\begin{minipage}[b]{.24\textwidth}
\centerline{
\begin{overpic}[width=1\textwidth]
{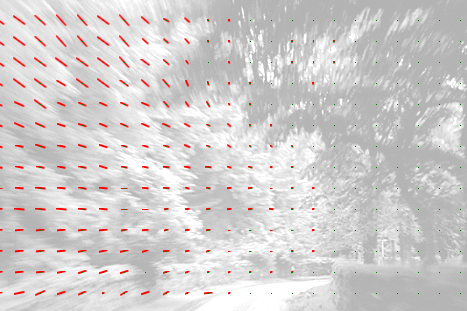}
\put(2,3){\footnotesize\color{black}{\bf (f) Motion flow of \cite{sun2015CNN} }}
\end{overpic}}
\end{minipage}
\begin{minipage}[b]{.24\textwidth}
\centerline{
\begin{overpic}[width=1\textwidth]
{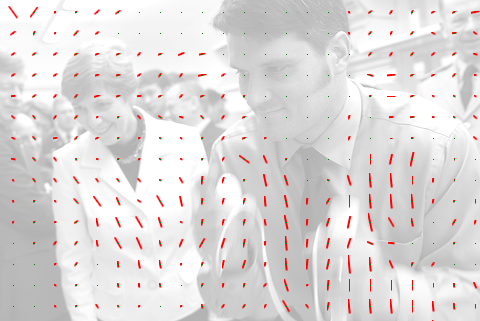}
\put(2,3){\footnotesize\color{black}{\bf (g) Motion flow of \cite{sun2015CNN} }}
\end{overpic}}
\end{minipage}
\begin{minipage}[b]{.24\textwidth}
\centerline{
\begin{overpic}[width=1\textwidth]
{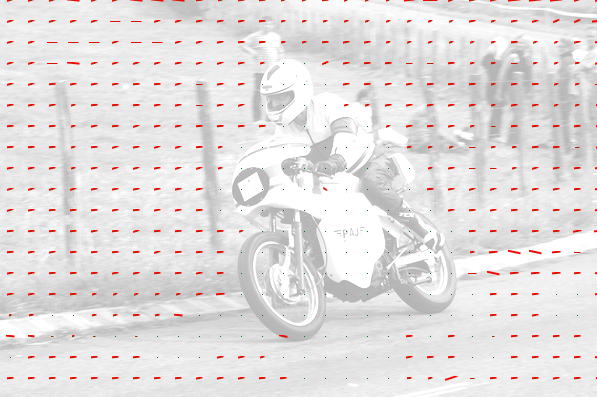}
\put(2,3){\footnotesize\color{black}{\bf (h) Motion flow of \cite{sun2015CNN} }}
\end{overpic}}
\end{minipage}
\begin{minipage}[b]{.24\textwidth}
\centerline{
\begin{overpic}[width=1\textwidth]
{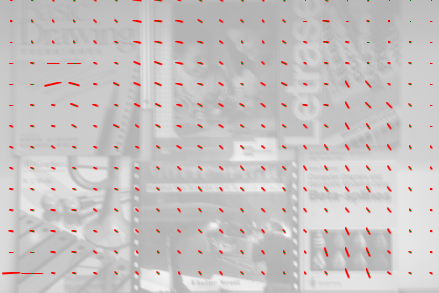}
\put(2,3){\footnotesize\color{black}{\bf (i) Our Motion flow}}
\end{overpic}}
\end{minipage}
\begin{minipage}[b]{.24\textwidth}
\centerline{
\begin{overpic}[width=1\textwidth]
{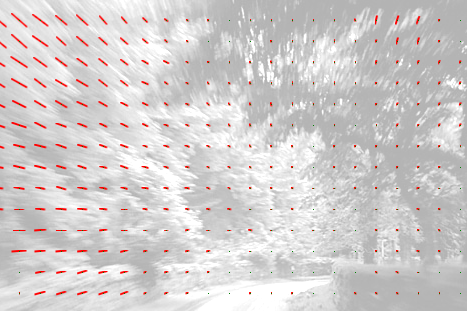}
\put(2,3){\footnotesize\color{black}{\bf (j) Our Motion flow}}
\end{overpic}}
\end{minipage}
\begin{minipage}[b]{.24\textwidth}
\centerline{
\begin{overpic}[width=1\textwidth]
{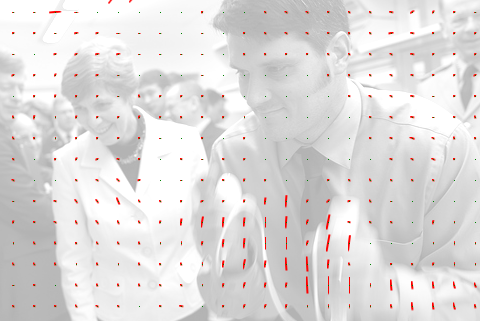}
\put(2,3){\footnotesize\color{black}{\bf (k) Our Motion flow}}
\end{overpic}}
\end{minipage}
\begin{minipage}[b]{.24\textwidth}
\centerline{
\begin{overpic}[width=1\textwidth]
{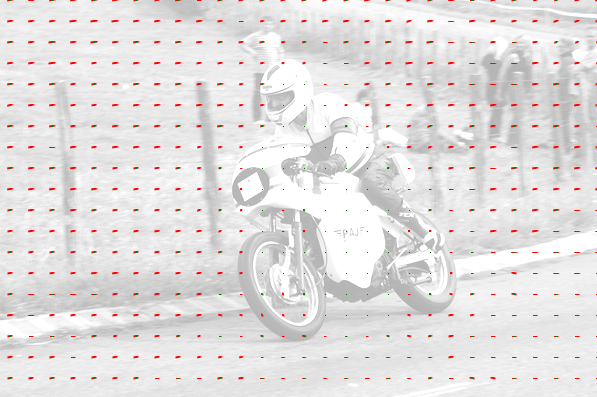}
\put(2,3){\footnotesize\color{black}{\bf (l) Our Motion flow}}
\end{overpic}}
\end{minipage}
\caption{Examples of motion flow estimation on real-world blurry images. From top to bottom: Blurry image $\B$, motion flow estimated by the patchCNN \cite{sun2015CNN}, and by our motion flow $\M$. Our results are more smooth and more accurate on moving objects.}
\label{fig:real_mf_est}
\end{figure*}

\subsection{Evaluation of Generalization Ability}
To evaluate the generalization ability of our approach on different images, we use the datasets based on the Microsoft COCO \cite{lin2014mscoco} (\ie MC-S and MC-M) to evaluate our model trained on the dataset based on BSD500 \cite{bsd500}.
Table \ref{tab:mse-mscoco} shows the evaluation and comparison with the ``patchCNN'' \cite{sun2015CNN}. The results demonstrate that our method stably produces high accuracy results on both datasets. This experiment suggests that the generalization ability of our approach is strong.

\begin{table}[htp]
\centering
\small
\caption{Evaluation of the generalization ability on datasets MC-S and MC-M. The best results are bold-faced.}
\label{tab:mse-mscoco}
\vspace{0.05cm}
\begin{tabular}{c|c|c |c c c}
\hline
    Dataset   & Metric & \!\!\! GT $\K$ \!\!\!&  \!\!\!patchCNN \!\!\! & \!\!\! noMRF \cite{sun2015CNN} \!\!\! & \!\!\! Ours \!\!\! \\ \hline
      &\!\!\! MSE\!\!\!   &  -- & 52.1234  & 60.9397 & \bf 7.8038 \\
\!\!\!MC-S\!\! &\!\!\! PSNR\!\!\!   &  22.620 &  20.172 & 20.217 & \bf 21.954 \\
      & \!\!\!SSIM\!\!\!  &  0.6953 &  0.5764 & 0.5772 & \bf 0.6641 \\ \hline
      &\!\!\! MSE \!\!\!  & -- & 22.4383 & 31.2754 & \bf 7.3405 \\
\!\!\!MC-M\!\! &\!\!\! PSNR \!\!\!  & 23.827 & 22.186 & 22.028 & \bf 23.227  \\
      & \!\!\!SSIM \!\!\! & 0.7620 & 0.6924 & 0.6839 & \bf 0.7402 \\ \hline
\end{tabular}
\end{table}

\subsection{Running-time Evaluation}
We conduct a running-time comparison with the relevant motion flow estimation methods \cite{sun2015CNN,kim2014segfree} by running motion flow estimation for 60 blurred images with sizes around $640\times 480$ on a PC with an NVIDIA GeForce 980 graphics card and Intel Core i7 CPU. For the method in \cite{kim2014segfree}, we quote its running-time from the paper. Note that both the method of Sun \etal and ours use the GPU to accelerate the computation. As shown in Table \ref{tab:time}, the method in \cite{kim2014segfree} takes very long time due to its iterative optimization scheme. Our method takes less than 10 seconds, which is more efficient than others. The patchCNN method \cite{sun2015CNN} takes more time because many post-processing steps are required.
\begin{table}[htp]
\centering
\caption{Running-time comparison.}
\vspace{0.05cm}
\label{tab:time}
\small
\begin{tabular}{c|cccc}
\hline
Method & \cite{kim2014segfree} & patchCNN \cite{sun2015CNN} &  noMRF \cite{sun2015CNN}  & Ours \\ \hline
Time (s)   & 1500 &  45.2 &  18.5 &\bf 8.4 \\ \hline
\end{tabular}
\end{table}

\subsection{Evaluation on Real-world Images} \label{sec:exp_real}
As the ground truth images of real-world blurry images are unavailable, we only present the visual evaluation and comparison against several state-of-the-art methods for spatially-varying blur removing. More results can be found in supplementary materials.

\par
\noindent \textbf{Results of motion flow estimation}
We first compare the proposed method with the method of Sun \etal \cite{sun2015CNN} on motion flow estimation. Four examples are shown in Figure \ref{fig:real_mf_est}. Since the method of Sun \etal performs on local patches, their motion flow components are often misestimated, especially when the blur pattern in a small local area is subtle or confusing, such as the areas with low illumination or textures. Thanks to the universal end-to-end mapping, our methods can generate more natural results with smooth flow and less clutters. Although we train our model on dataset with only smoothly varying motion flow, compared with \cite{sun2015CNN}, our method can obtain better results on images with moving object.

\begin{figure}[htp]
\centering
\begin{minipage}[b]{.15\textwidth}
\centerline{
\begin{overpic}[trim=0 0 5 5, clip, width=1\textwidth]
{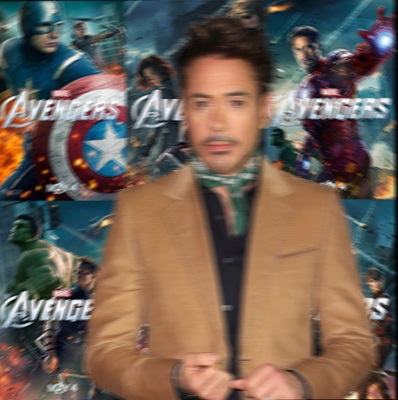}
\put(2,3){\footnotesize\color{white}{\bf (a) Blurry image}}
\end{overpic}}
\end{minipage}
\begin{minipage}[b]{.15\textwidth}
\centerline{
\begin{overpic}[trim=0 0 5 5, clip, width=1\textwidth]
{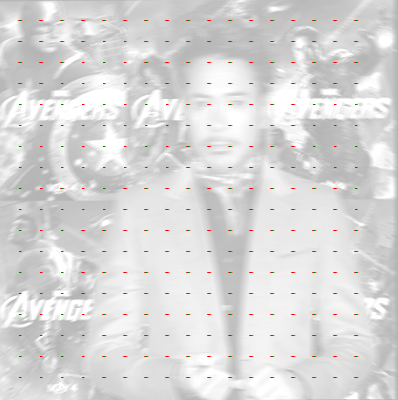}
\put(2,3){\footnotesize\color{black}{\bf (b) \cite{kim2014segfree} }}
\end{overpic}}
\end{minipage}
\begin{minipage}[b]{.15\textwidth}
\centerline{
\begin{overpic}[trim=0 0 5 5, clip, width=1\textwidth]
{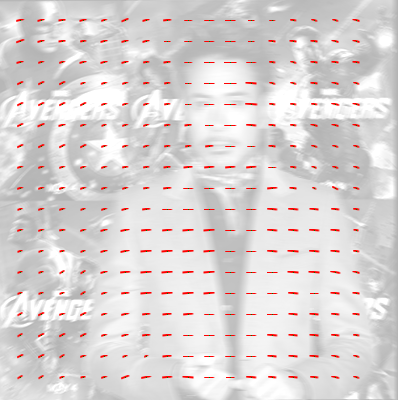}
\put(2,3){\footnotesize\color{black}{\bf (c) Ours }}
\end{overpic}}
\end{minipage}
\begin{minipage}[b]{.15\textwidth}
\centerline{
\begin{overpic}[trim=0 0 5 5, clip, width=1\textwidth]
{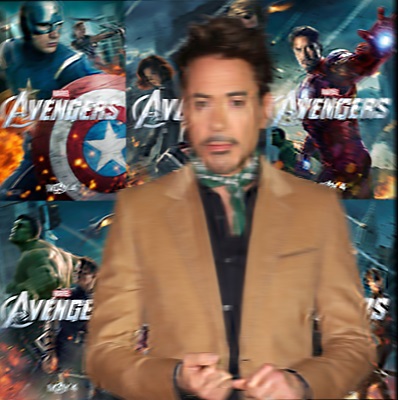}
\put(2,3){\footnotesize\color{white}{\bf (d) \cite{sun2015CNN}}}
\end{overpic}}
\end{minipage}
\begin{minipage}[b]{.15\textwidth}
\centerline{
\begin{overpic}[trim=0 0 5 5, clip, width=1\textwidth]
{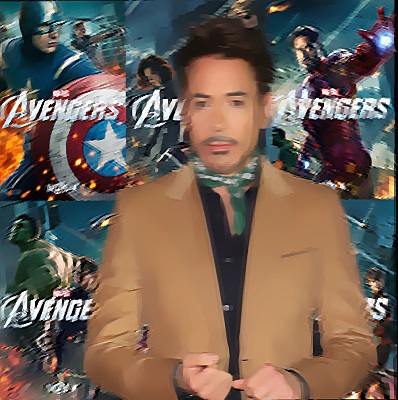}
\put(2,3){\footnotesize\color{white}{\bf (e) \cite{kim2014segfree}}}
\end{overpic}}
\end{minipage}
\begin{minipage}[b]{.15\textwidth}
\centerline{
\begin{overpic}[trim=0 0 5 5, clip, width=1\textwidth]
{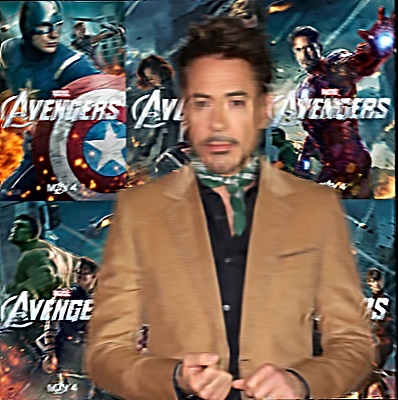}
\put(2,3){\footnotesize\color{white}{\bf (f) Ours}}
\end{overpic}}
\end{minipage}
\caption{Comparison with the method of Kim and Lee \cite{kim2014segfree}.}
\label{fig:com_kim}
\vspace{-0.1cm}
\end{figure}

\noindent \textbf{Comparison with the method in \cite{kim2014segfree}}
Kim \etal \cite{kim2014segfree} use the similar heterogeneous motion blur model as ours and also estimate motion flow for deblurring. As their code is unavailable, we directly perform a comparison on their real-world data. Figure \ref{fig:com_kim} shows the results on an example. Compared with the results of Kim and Lee \cite{kim2014segfree}, our motion flow more accurately reflects the complex blur pattern, and our recovered image contains more details and less artifacts.

\begin{figure*}[htp]
\centering
\subfigure[Blurry image]{
\begin{minipage}[b]{.245\textwidth}
\centerline{
\begin{overpic}[width=1\textwidth]
{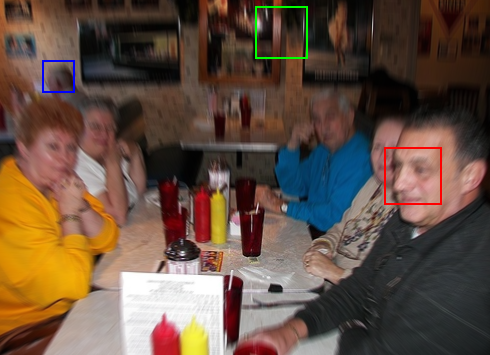}
\put(-1,2){ \includegraphics[width=0.15\textwidth]{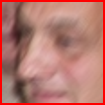}}
\put(16,2){ \includegraphics[width=0.15\textwidth]{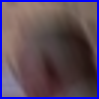}}
\put(33,2){ \includegraphics[width=0.15\textwidth]{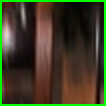}}
\end{overpic}}
\end{minipage}
}
\hspace{-0.3cm}
\subfigure[{Whyte \etal \cite{whyte2012non}}]{
\begin{minipage}[b]{.245\textwidth}
\centerline{
\begin{overpic}[width=1\textwidth]
{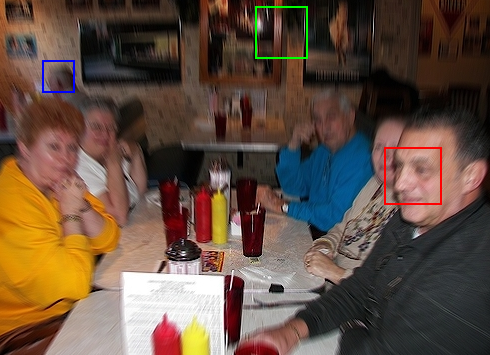}
\put(-1,2){ \includegraphics[width=0.15\textwidth]{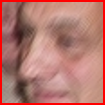}}
\put(16,2){ \includegraphics[width=0.15\textwidth]{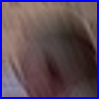}}
\put(33,2){ \includegraphics[width=0.15\textwidth]{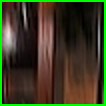}}
\end{overpic}}
\end{minipage}
}
\hspace{-0.3cm}
\subfigure[Sun \etal \cite{sun2015CNN}]{
\begin{minipage}[b]{.245\textwidth}
\centerline{
\begin{overpic}[width=1\textwidth]
{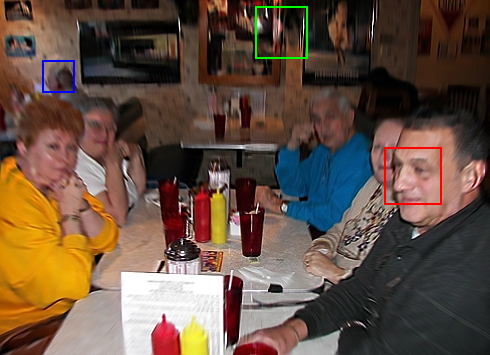}
\put(-1,2){ \includegraphics[width=0.15\textwidth]{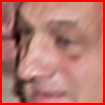}}
\put(16,2){ \includegraphics[width=0.15\textwidth]{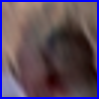}}
\put(33,2){ \includegraphics[width=0.15\textwidth]{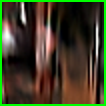}}
\end{overpic}}
\end{minipage}
}
\hspace{-0.3cm}
\subfigure[Ours]{
\begin{minipage}[b]{.245\textwidth}
\centerline{
\begin{overpic}[width=1\textwidth]
{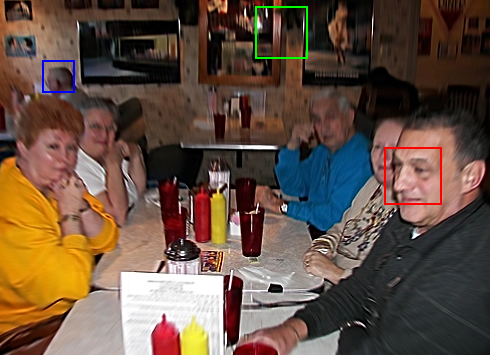}
\put(-1,2){ \includegraphics[width=0.15\textwidth]{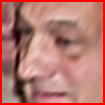}}
\put(16,2){ \includegraphics[width=0.15\textwidth]{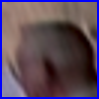}}
\put(33,2){ \includegraphics[width=0.15\textwidth]{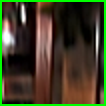}}
\end{overpic}}
\end{minipage}
}
\caption{Deblurring results on an image with camera motion blur.}
\label{fig:cam_m}
\vspace{-0.1cm}
\end{figure*}

\begin{figure*}[!t]
\centering
\subfigure[Blurry image]{
\begin{minipage}[b]{.195\textwidth}
\centerline{
\begin{overpic}[trim=25 10 5 20, clip, width=1\textwidth]
{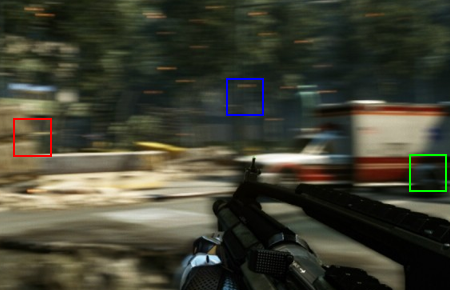}
\put(3,2.6){ \includegraphics[width=0.13\textwidth]{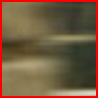}}
\put(22.5,2.6){ \includegraphics[width=0.127\textwidth]{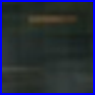}}
\put(83,2.6){ \includegraphics[width=0.124\textwidth]{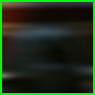}}
\end{overpic}}
\end{minipage}
}
\hspace{-0.3cm}
\subfigure[Whyte \etal \cite{whyte2012non}]{
\begin{minipage}[b]{.195\textwidth}
\centerline{
\begin{overpic}[trim=25 10 5 20, clip, width=1\textwidth]
{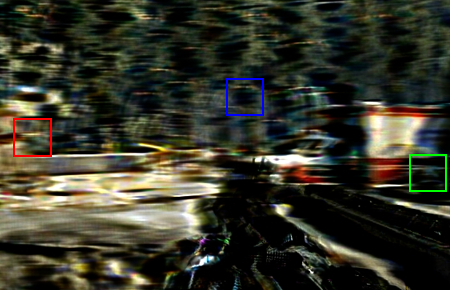}
\put(3,2.6){ \includegraphics[width=0.13\textwidth]{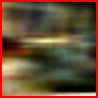}}
\put(22.5,2.6){ \includegraphics[width=0.127\textwidth]{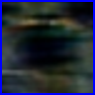}}
\put(83,2.6){ \includegraphics[width=0.124\textwidth]{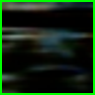}}
\end{overpic}}
\end{minipage}
}
\hspace{-0.3cm}
\subfigure[Kim and Lee \cite{kim2014segfree}]{
\begin{minipage}[b]{.195\textwidth}
\centerline{
\begin{overpic}[trim=25 10 5 20, clip, width=1\textwidth]
{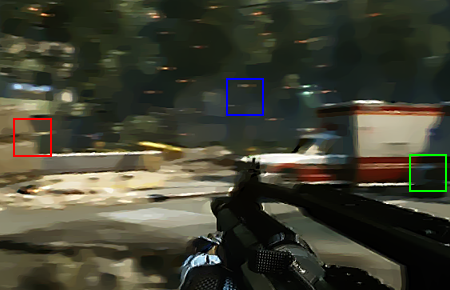}
\put(3,2.6){ \includegraphics[width=0.13\textwidth]{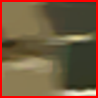}}
\put(22.5,2.6){ \includegraphics[width=0.127\textwidth]{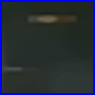}}
\put(83,2.6){ \includegraphics[width=0.124\textwidth]{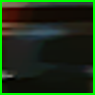}}
\end{overpic}}
\end{minipage}
}
\hspace{-0.3cm}
\subfigure[Sun \etal \cite{sun2015CNN}]{
\begin{minipage}[b]{.195\textwidth}
\centerline{
\begin{overpic}[trim=25 10 5 20, clip, width=1\textwidth]
{results/back_m/kim_016_0.png}
\put(3,2.6){ \includegraphics[width=0.13\textwidth]{results/back_m/kim_016_1.png}}
\put(22.5,2.6){ \includegraphics[width=0.127\textwidth]{results/back_m/kim_016_2.png}}
\put(83,2.6){ \includegraphics[width=0.124\textwidth]{results/back_m/kim_016_3.png}}
\end{overpic}}
\end{minipage}
}
\hspace{-0.3cm}
\subfigure[Ours]{
\begin{minipage}[b]{.195\textwidth}
\centerline{
\begin{overpic}[trim=25 10 5 20, clip, width=1\textwidth]
{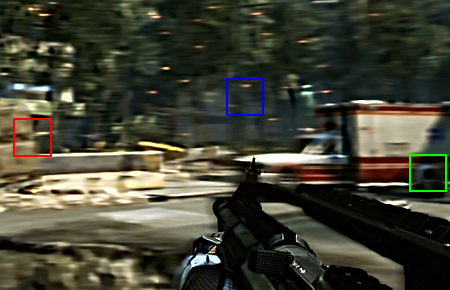}
\put(3,2.6){ \includegraphics[width=0.13\textwidth]{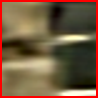}}
\put(22.5,2.6){ \includegraphics[width=0.127\textwidth]{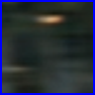}}
\put(83,2.6){ \includegraphics[width=0.124\textwidth]{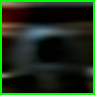}}
\end{overpic}}
\end{minipage}
}
\caption{Deblurring results on an non-uniform blur image with strong blur on background.}
\label{fig:back_m}
\vspace{-0.1cm}
\end{figure*}

\begin{figure*}[!t]
\centering
\subfigure[Blurry image]{
\begin{minipage}[b]{.245\textwidth}
\centerline{
\begin{overpic}[width=1\textwidth]
{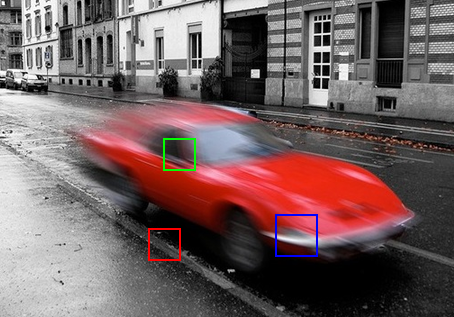}
\put(-2,46.7){ \includegraphics[width=0.23\textwidth]{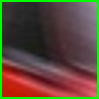}}
\put(21.1,46.7){ \includegraphics[width=0.23\textwidth]{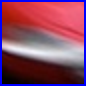}}
\put(-2,0){ \includegraphics[width=0.23\textwidth]{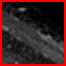}}
\end{overpic}}
\end{minipage}
}
\hspace{-0.3cm}
\subfigure[Pan \etal \cite{pan2016soft}]{
\begin{minipage}[b]{.245\textwidth}
\centerline{
\begin{overpic}[width=1\textwidth]
{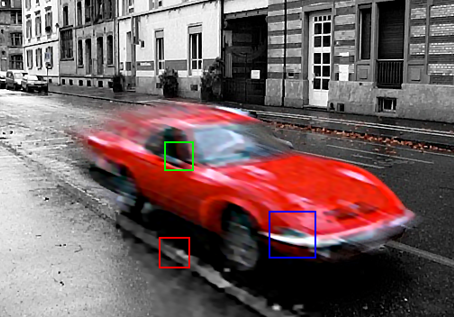}
\put(-2,46.7){ \includegraphics[width=0.23\textwidth]{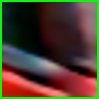}}
\put(21.1,46.7){ \includegraphics[width=0.23\textwidth]{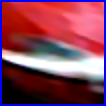}}
\put(-2,0){ \includegraphics[width=0.23\textwidth]{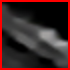}}
\end{overpic}}
\end{minipage}
}
\hspace{-0.3cm}
\subfigure[Sun \etal \cite{sun2015CNN}]{
\begin{minipage}[b]{.245\textwidth}
\centerline{
\begin{overpic}[width=1\textwidth]
{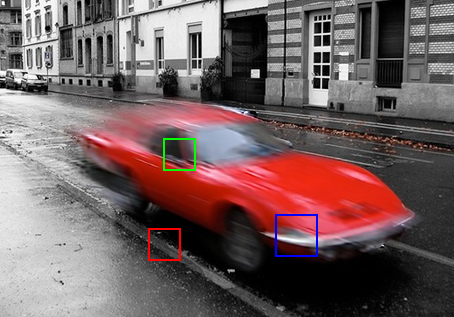}
\put(-2,46.7){ \includegraphics[width=0.23\textwidth]{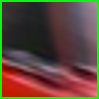}}
\put(21.1,46.7){ \includegraphics[width=0.23\textwidth]{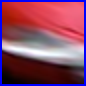}}
\put(-2,0){ \includegraphics[width=0.23\textwidth]{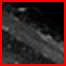}}
\end{overpic}}
\end{minipage}
}
\hspace{-0.3cm}
\subfigure[Ours]{
\begin{minipage}[b]{.245\textwidth}
\centerline{
\begin{overpic}[width=1\textwidth]
{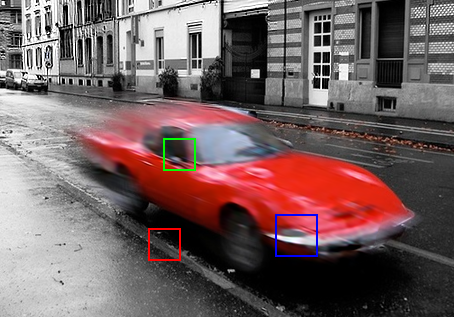}
\put(-2,46.7){ \includegraphics[width=0.23\textwidth]{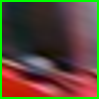}}
\put(21.1,46.7){ \includegraphics[width=0.23\textwidth]{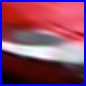}}
\put(-2,0){ \includegraphics[width=0.23\textwidth]{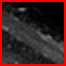}}
\end{overpic}}
\end{minipage}
}
\caption{Deblurring results on an image with large scale motion blur caused by moving object.}
\label{fig:redcar}
\vspace{-0.2cm}
\end{figure*}

\noindent \textbf{Images with camera motion blur}
Figure \ref{fig:cam_m} shows an example containing blur mainly caused by the camera motion. The deblurred image generated by the non-uniform camera shake deblurring method \cite{whyte2012non} suffers from heavy blur because its model ignores the blur caused by large forward motion. Compared with the result of Sun \etal \cite{sun2015CNN}, our method produces a sharper result with more details and less artifacts.

\noindent \textbf{Images with object motion blur} We evaluate our method on the images containing object motion blur. In Figure \ref{fig:back_m}, the result of Whyte \etal \cite{whyte2012non} contains heavy ringing artifacts due to the object motion. Our method can handle the strong blur in the background and generate a more natural image.
We further compare with the segmentation-based deblurring method of Pan \etal \cite{pan2016soft} on an image with large scale blur caused by moving object on static background. As shown in Figure \ref{fig:redcar}, the result of Sun \etal \cite{sun2015CNN} is oversmooth due to the underestimate of motion flow. In the result of Pan \etal \cite{pan2016soft}, some details are lost due to the segmentation error. Our proposed method can recover the details on blurred moving foreground and keep the sharp background as original.

\section{Conclusion}
In this paper, we proposed a flexible and efficient deep learning based method for estimating and removing the heterogeneous motion blur. By representing the heterogeneous motion blur as pixel-wise linear motion blur, the proposed method uses a FCN to estimate the a dense motion flow map for blur removal. Moreover, we automatically generate training data with simulated motion flow maps for training the FCN.
Experimental results on both synthetic and real-world data show the excellence of the proposed method.

{\small
\bibliographystyle{ieee}
\bibliography{blur2motion}
}

\end{document}